\documentclass[10pt,journal,compsoc]{IEEEtran}

\usepackage{amsmath,amssymb,amsfonts}
\usepackage{booktabs}
\usepackage{graphicx}
\usepackage[nocompress]{cite}
\usepackage{url}
\usepackage[table]{xcolor}
\usepackage{array}
\usepackage{stfloats}
\usepackage{textcomp}
\usepackage{multirow}

\usepackage{amsmath,amsfonts,bm}

\def\eqref#1{equation~\ref{#1}}

\def\1{\bm{1}}

\DeclareMathAlphabet{\mathsfit}{\encodingdefault}{\sfdefault}{m}{sl}
\SetMathAlphabet{\mathsfit}{bold}{\encodingdefault}{\sfdefault}{bx}{n}

\newcommand{\ucfltx}{66.8}
\newcommand{\ucfdrivelaw}{68.3}
\newcommand{\ucfreworld}{71.7}
\newcommand{\ucfreworldfull}{80.2}
\newlength{\ablationw}
\newlength{\inferw}
\newlength{\ablationshortw}
\newlength{\fidtabw}
\AtBeginDocument{%
  \setlength{\ablationw}{0.92\columnwidth}%
  \setlength{\inferw}{0.95\columnwidth}%
  \setlength{\ablationshortw}{0.72\columnwidth}%
  \setlength{\fidtabw}{0.88\columnwidth}%
  \setcounter{topnumber}{2}%
  \setcounter{bottomnumber}{2}%
  \setcounter{totalnumber}{4}%
}
\usepackage{hyperref}
\hypersetup{colorlinks=true,linkcolor=blue,citecolor=blue,urlcolor=blue}

\usepackage{etoolbox}
\makeatletter
\def\@IEEEBIOphotowidth{1.0in}
\def\@IEEEBIOphotodepth{1.25in}
\def\@IEEEBIOhangwidth{1.14in}
\def\@IEEEBIOhangdepth{1.25in}
\def\@IEEEBIOskipN{0.45in}
\patchcmd{\IEEEbiography}%
  {\vskip \@IEEEBIOskipN plus 1fil minus 0\baselineskip}%
  {\vskip \@IEEEBIOskipN}%
  {}{\typeout{Warning: failed to patch IEEEbiography skip}}
\makeatother
\newcommand{\biosphoto}[1]{%
  \includegraphics[width=1.0in,clip,keepaspectratio]{#1}}

\begin{document}

\title{ReWorld: Representation Learning for World Action Models}

\author{Tianze~Xia,
        Lijun~Zhou,
        Kaixin~Xiong,
        Jingfeng~Yao,
        Zhenxin~Zhu,
        Haiyang~Sun,
        Bing~Wang,
        Guang~Chen,
        Wenyu~Liu,~\IEEEmembership{Senior~Member,~IEEE,}
        Hangjun~Ye,
        and~Xinggang~Wang,~\IEEEmembership{Senior~Member,~IEEE}%
\IEEEcompsocitemizethanks{\IEEEcompsocthanksitem Tianze Xia, Jingfeng Yao, Wenyu Liu, and Xinggang Wang are with Huazhong University of Science and Technology, Wuhan, China.\IEEEcompsocthanksitem Tianze Xia, Lijun Zhou, Kaixin Xiong, Zhenxin Zhu, Haiyang Sun, Bing Wang, Guang Chen, and Hangjun Ye are with Xiaomi EV, China.\IEEEcompsocthanksitem Tianze Xia and Lijun Zhou contributed equally to this work. Lijun Zhou is the project lead.\IEEEcompsocthanksitem Corresponding author: Xinggang Wang (E-mail: \url{xgwang@hust.edu.cn}).}}

\markboth{SUBMITTED TO IEEE Transactions on Pattern Analysis and Machine Intelligence}%
{Xia \MakeLowercase{\textit{et al.}}: ReWorld: Representation Learning for World Action Models}

\IEEEtitleabstractindextext{%
\begin{abstract}
World Action Models (WAMs) unify future environment prediction with action generation for autonomous driving, yet existing approaches optimize only the final outputs, leaving intermediate representations as incidental byproducts. We present \textbf{ReWorld}, the first representation learning framework specifically designed for autonomous-driving WAMs. ReWorld explicitly optimizes the latent world-to-action pathway through three complementary mechanisms. First, it imposes future-predictive supervision on intermediate Video DiT states to encode temporal scene dynamics, enabling self-guided sampling and a roughly twofold convergence speedup. Second, it aligns Action DiT states with their attended video readouts so that the retrieved world information is retained in the representations used for planning. Third, it shapes the action space using geometrically close yet low-scoring hard negatives to separate the expert trajectory from nearby unsafe alternatives. ReWorld constructs supervision entirely from the WAM's own generation targets and attended features, requiring no external encoders or teacher models and introducing only 0.3\% additional per-step training cost. Experiments show that ReWorld reduces FVD from 81.3 to 61.9 on nuScenes, improves closed-loop PDMS from 89.1 to 90.4 on NAVSIM without reinforcement learning or test-time scoring, and increases frozen linear-probe accuracy from 68.3\% to 80.2\% on UCF-101 action recognition. These results indicate that explicitly optimized representations are central to translating world knowledge into planning capability in WAMs. Code is available at \url{https://github.com/xiaomi-research/ReWorld}.
\end{abstract}

}

\maketitle
\IEEEdisplaynontitleabstractindextext

\begin{figure*}[t]
  \centering
  \includegraphics[width=\linewidth]{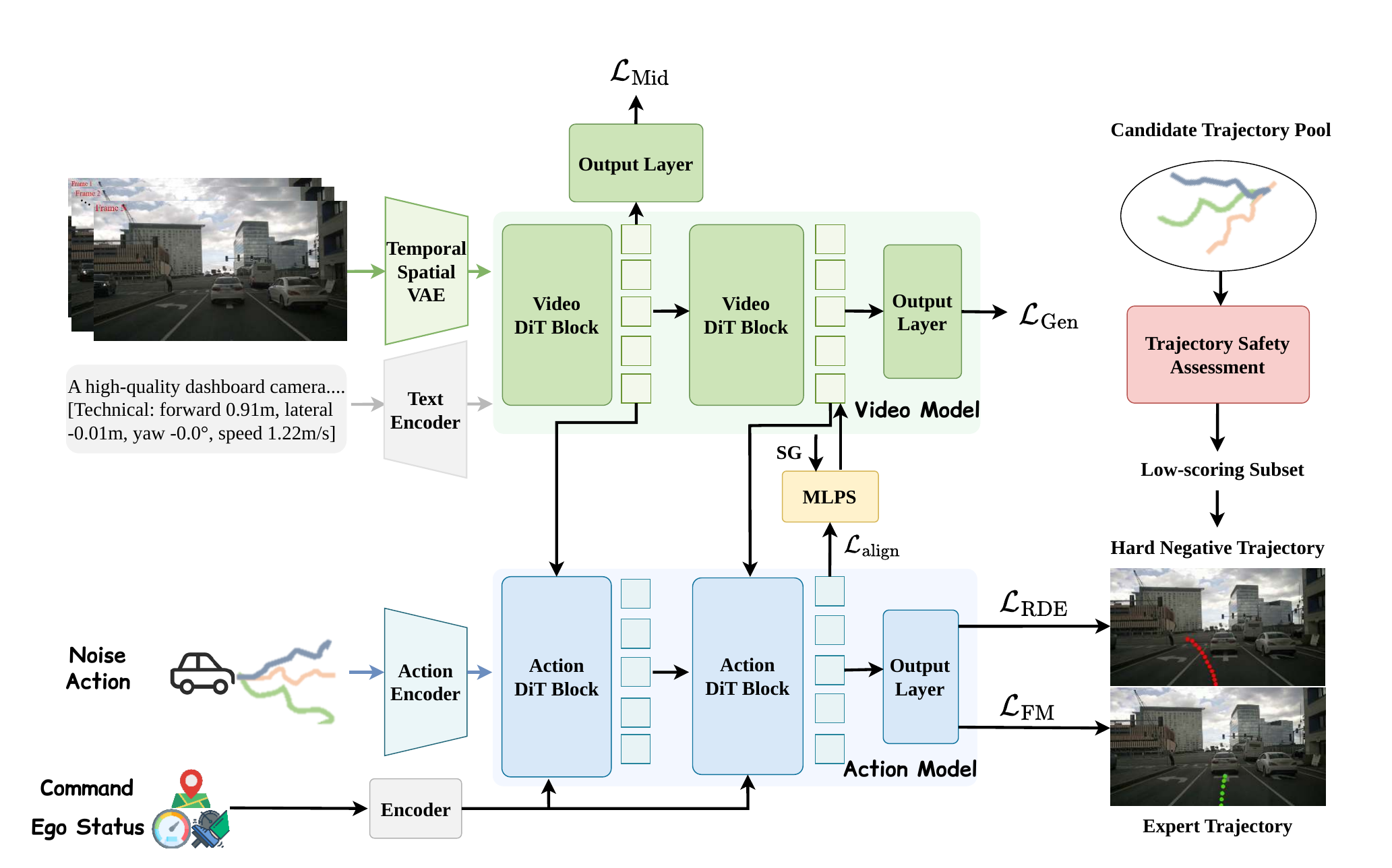}
  \caption{\textbf{Overview of ReWorld.} A Video DiT learns a latent representation of future scene evolution, whose mid-denoising states $\mathcal{F}$ condition an Action DiT for trajectory generation. ReWorld explicitly optimizes this world-to-action representation pathway in three stages. Stage~1 makes intermediate video states future-predictive through $\mathcal{L}_{\mathrm{Mid}}$ and enables self-guided video sampling. Stage~2 freezes the Video DiT and aligns post-cross-attention action states with their attended video readouts through $\mathcal{L}_{\mathrm{align}}$. Stage~3 jointly fine-tunes both branches with $\mathcal{L}_{\mathrm{RDE}}$, using geometrically close but low-scoring trajectories to shape behavior-sensitive action representations.}
  \label{fig:framework}
\end{figure*}

\IEEEraisesectionheading{\section{Introduction}\label{sec:introduction}}

\IEEEPARstart{W}{orld} models have emerged as an important direction in autonomous driving by learning how driving scenes evolve over time and providing dynamic priors beyond instantaneous perception~\cite{li2024enhancing,wang2024driving,li2025drivevla,zhang2025epona}. In particular, video-generation-based world models learn scene dynamics and physical regularities directly from large-scale driving videos through future pixel prediction~\cite{gao2023magicdrive,gao2024vista,hu2023gaia,li2024drivingdiffusion,wang2024drivedreamer,zhao2025drivedreamer}. However, video prediction is not the ultimate goal of a decision-oriented system. World Action Models (WAMs) go beyond visual forecasting by coupling future world prediction with action generation, allowing knowledge learned from video to support decision-making directly~\cite{chen2025drivinggpt,zhang2025epona,li2025drivevla,zheng2024genad,li2025omninwm,liu2026driveva}.

A central challenge for WAMs is transferring world knowledge from video generation to planning. Simulator-based methods and auxiliary prediction objectives affect planning only indirectly, while joint generation--planning models often preserve separate output streams~\cite{zhang2025epona,chen2025drivinggpt}. DriveLaW~\cite{xia2026drivelaw} introduces a chained formulation that conditions an Action DiT directly on intermediate Video DiT features cached at the initial reverse-flow step (Fig.~\ref{fig:framework}). This latent connection turns the video generator from a visual renderer into a source of planning representations.

Yet latent access alone does not ensure effective representations. With standard output-level objectives, intermediate Video DiT states are supervised only through the final denoising output, and Action DiT states only through the final trajectory prediction. Consequently, video states may lack explicit future-predictive structure, while action states may fail to retain the world information retrieved through cross-attention. Imitation learning further provides limited supervision for distinguishing expert actions from geometrically similar yet unsafe alternatives. We term this limitation the \emph{representation bottleneck of WAMs}: the intermediate representations connecting world prediction to action generation are not explicitly optimized to be future-predictive, cross-modally grounded, or sensitive to closed-loop behavior quality.

We present \textbf{ReWorld}, the first representation learning framework specifically designed for autonomous-driving WAMs. ReWorld addresses this bottleneck through a progressive curriculum that optimizes representation formation, cross-modal transfer, and decision-oriented shaping. First, auxiliary prediction heads make intermediate Video DiT states explicitly future-predictive. The resulting shallow-to-deep prediction hierarchy further enables inference-time self-guidance (Fig.~\ref{fig:infer}) and a roughly twofold convergence speedup from scratch. Second, ReWorld aligns Action DiT states with the video readouts retrieved through cross-attention, grounding action representations in world information while keeping the video branch frozen. Third, it jointly fine-tunes both branches with geometrically close yet low-scoring hard negatives, making the action space sensitive to unsafe behaviors near the expert trajectory.

Representation learning methods developed for image diffusion provide useful precedents~\cite{yu2024representation,jiang2025no}, but do not transfer directly to long-horizon driving video. External image or video encoders may introduce semantic targets misaligned with the generator's flow-matching dynamics, while teacher branches add substantial training cost. ReWorld instead derives supervision entirely from the WAM's own generation targets, attended video readouts, and trajectory candidates, requiring no external representation encoders or teacher models and introducing only 0.3\% additional per-step Video DiT training cost.

We evaluate ReWorld on nuScenes~\cite{caesar2020nuscenes} and NAVSIM~\cite{dauner2024navsim}, and analyze frozen Video DiT representations on UCF-101. ReWorld reduces FVD from 81.3 to 61.9 on nuScenes, improves closed-loop PDMS from 89.1 to 90.4 on NAVSIM without reinforcement learning or test-time scoring, and increases frozen linear-probe accuracy from 68.3\% to 80.2\% on UCF-101 action recognition. It also converges roughly twice as fast from scratch and consistently outperforms existing diffusion representation-learning methods under a unified training protocol. These results indicate that explicitly optimized representations are central to translating world knowledge into planning capability in WAMs.

The main contributions are summarized as follows:

(1) We present ReWorld, the first representation learning framework specifically designed to optimize the latent world-to-action pathway in autonomous-driving WAMs.

(2) We introduce a progressive curriculum that makes Video DiT states future-predictive, grounds Action DiT states in attended video information, and shapes action representations with geometrically close yet low-scoring hard negatives. These objectives require no external encoders or teacher models and add only 0.3\% per-step Video DiT training cost.

(3) We systematically evaluate ReWorld at the output, representation, and decision levels. ReWorld improves video generation, planning, and frozen representation quality, accelerates convergence, and consistently outperforms existing diffusion representation learning methods in controlled experiments.

\section{Related Work}
\label{sec:related}

\subsection{World Models for Driving Video}

Video-generation-based world models are widely used for scene generation, data augmentation, and closed-loop simulation~\cite{hu2023gaia,wang2024drivedreamer,gao2023magicdrive,gao2024vista,li2024drivingdiffusion,wen2024panacea,zhao2025drivedreamer,russell2025gaia}. World models more broadly aim to internalize physical structure and dynamics into predictive representations~\cite{brooks2024video,bruce2024genie,assran2025v}. Modeling paradigms have evolved from autoregressive token predictors such as DrivingGPT~\cite{chen2025drivinggpt} to high-fidelity diffusion generators such as MiLA~\cite{wang2025mila}, while OccWorld~\cite{zheng2024occworld}, OccSora~\cite{wang2024occsora}, UniScene~\cite{li2025uniscene}, and Genesis~\cite{guo2025genesis} strengthen 3D structure and cross-modal consistency. Another line treats video world models as simulators for evaluation and policy learning, including HUGSIM~\cite{zhou2024hugsim}, RAD~\cite{gao2025rad}, ReSim~\cite{yang2025resim}, ReconDreamer-RL~\cite{ni2025recondreamer}, and OmniNWM~\cite{li2025omninwm}.

These advances improve rendering quality, controllability, and simulation utility, yet most work still emphasizes future video prediction itself. How video generators can obtain better \emph{internal} representations---for training efficiency, temporal coherence, and downstream planning---remains comparatively underexplored. For planning-facing WAMs, representation quality is especially important: mid-denoising video states must carry future-predictive structure rather than only photorealistic appearance.

\begin{figure*}[t]
  \centering
  \includegraphics[width=\linewidth]{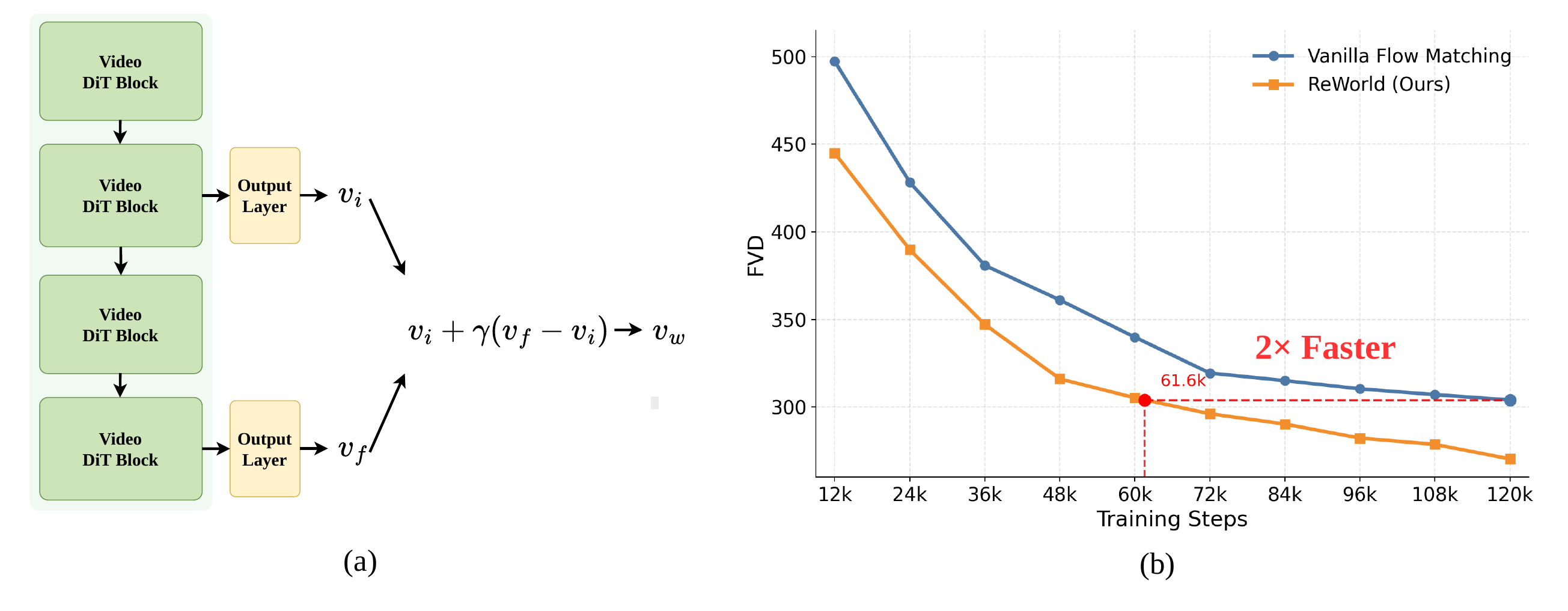}
  \caption{\textbf{Intermediate-supervised inference and accelerated convergence of ReWorld.} (a)~ReWorld uses the discrepancy between the intermediate prediction $v_i$ and the final prediction $v_f$ to construct the self-guided velocity $v_w$. (b)~ReWorld reaches a comparable validation level using approximately half the optimization steps of vanilla flow matching, corresponding to an approximately $2\times$ convergence acceleration. Neither training scheme uses an external representation encoder or teacher model.}
  \label{fig:infer}
\end{figure*}

\subsection{World Action Models}

As world models move closer to decision-making, research is shifting from scene simulation toward World Action Models that predict future evolution under action conditioning~\cite{chen2025drivinggpt,zhang2025epona,li2025drivevla,bartoccioni2025vavim,xia2026drivelaw,liu2026driveva,wang2026latent}. Existing designs differ mainly in how tightly generation and planning share internal state. Shared-backbone co-generators with separate heads (DrivingGPT~\cite{chen2025drivinggpt}, Epona~\cite{zhang2025epona}, PWM~\cite{zhao2026forecasting}, DriveDreamer-policy~\cite{zhou2026drivedreamer}, DriveVA~\cite{liu2026driveva}) and sequential ``video then policy'' pipelines (GenAD~\cite{zheng2024genad}, OmniNWM~\cite{li2025omninwm}) still often keep imagined futures and planned actions as parallel streams. Related unified generators further include $\pi_0$-like mixture-of-transformers designs~\cite{black2024pi_0,liang2024mixture,bartoccioni2025vavim,li2025drivevla}, visual CoT unification~\cite{zeng2025futuresightdrive}, and decision-oriented world models~\cite{wang2024driving,li2024enhancing,li2025end}. In contrast, cascading a video generator and a planner so that mid-level video latents condition the Action DiT---as in DriveLaW~\cite{xia2026drivelaw}, inspired by Genie Envisioner-style chaining~\cite{liao2025genie}---treats the generator as a world-state provider rather than only a renderer.

Chained latent architectures provide a direct path for transferring video-generation priors into planning. ReWorld focuses on a complementary question: how the intermediate states along this path should be explicitly learned. It therefore studies representation formation in the Video DiT, representation transfer through video--action cross-attention, and behavior-oriented shaping in the Action DiT.

\subsection{Representation Learning for Generative Models}

Representation learning for diffusion generators falls into three broad lines. The first reshapes the latent space on which generation operates, from classical LDM~\cite{rombach2022high} in a VAE latent space~\cite{kingma2013auto} toward semantically richer autoencoders such as RAE~\cite{zheng2025rae}, SVG~\cite{shi2025latent}, VA-VAE~\cite{yao2025reconstruction}, VFM-VAE~\cite{bi2026vision}, AlignTok~\cite{chen2025aligning}, and FAE~\cite{gao2026one}. The second optimizes intermediate DiT/SiT~\cite{peebles2023scalable,ma2024sit} features during training: REPA~\cite{yu2024representation} aligns them with external encoders, with extensions in REPA-E~\cite{leng2025repa}, U-REPA~\cite{tian2026u}, and iREPA~\cite{singh2025matters}; SRA~\cite{jiang2025no} replaces teachers with self-alignment, while DiverseDiT~\cite{yang2026diversedit}, ReDi~\cite{kouzelis2026boosting}, SFD~\cite{pan2026semantics}, and REG~\cite{wu2026representation} explore diversity and token-level objectives. The third uses latent predictions at inference to refine sampling, as in Latent Forcing~\cite{baade2026latent}.

Most of this literature targets image generation. Transfer to long-horizon driving video remains underexplored, and gains from external teachers are often inconsistent in this regime: encoders pretrained on static images or short clips may supply semantic priors that are poorly matched to multi-second ego and agent dynamics, while self-alignment methods designed for images may not stress the temporal horizon that driving requires. ReWorld instead studies representation objectives that serve both generation and planning without an external teacher, evaluated under a protocol that stresses long-horizon temporal modeling and planning-oriented control.

\section{Method}
\label{sec:method}

\subsection{Preliminaries}
\label{sec:chained}

ReWorld is instantiated on a chained latent World Action Model (WAM), in which a Video DiT models future scene evolution and an Action DiT generates ego trajectories conditioned on the Video DiT's mid-denoising states. We follow the DriveLaW architecture~\cite{xia2026drivelaw} for this latent world-to-action interface. Unlike parallel designs that connect generation and planning only through decoded frames or separate policy heads, the chained design conditions the Action DiT~\cite{peebles2023scalable} directly on mid-denoising video features $\mathcal{F}$. These features are extracted at the initial reverse-flow step, corresponding to the maximum-noise endpoint of video flow time and the first discrete video denoising step. They provide the planner with an abstract scene representation shaped by the video-generation prior~\cite{liao2025genie}.

Concretely, the chained WAM consists of a 2B Video DiT~\cite{peebles2023scalable} and a 133M Action DiT (Fig.~\ref{fig:framework}). The Video DiT models future scene evolution from highly compressed causal VAE latents and employs hybrid late-stage pixel decoding~\cite{hacohen2024ltx}. Following diffusion-based planners~\cite{zheng2025diffusion,liao2025diffusiondrive,li2025recogdrive}, the Action DiT predicts ego trajectories directly from mid-denoising video features rather than from a separate perception or policy head. At the first discrete video denoising step, we retain the block features
$\mathcal{F}=\{f^{(b)}\}_{b=1}^{B}$ within the current forward pass and condition the planner on them. Thus, imagined scene evolution and planned motion are connected through a shared latent interface. Here, ``cached'' means that these activations are reused across action flow-matching steps, rather than precomputed or detached offline. The standard WAM is pretrained with a motion-first progressive schedule under output-level supervision~\cite{xia2026drivelaw}.

Given a driving clip $x$, ego kinematics $s_{\leq 0}$, and navigation command $g$, the spatiotemporal VAE~\cite{hacohen2024ltx} encodes the clip into a compact latent representation $z_0=E(x)$. Ego motion and navigation intent for the video branch are converted into a motion-conditioned natural-language prompt and encoded by a frozen T5 text encoder, producing the condition embedding $c^v$. Under the rectified-flow parameterization~\cite{liu2022flow}, we independently sample the video flow time $t_v\sim\mathcal{U}(0,1)$ and construct
\begin{equation}
z_{t_v}=(1-t_v)z_0+t_v\epsilon_z,
\qquad \epsilon_z\sim\mathcal{N}(0,I).
\label{eq:video-noise}
\end{equation}
The Video DiT velocity field $v_\theta^z$ is trained with
\begin{equation}
\mathcal{L}_{\mathrm{Gen}}
=
\mathbb{E}_{z_0,t_v,\epsilon_z}
\left[
\left\|
v_\theta^z(z_{t_v},t_v,c^v)-(\epsilon_z-z_0)
\right\|_2^2
\right].
\label{eq:diff}
\end{equation}

The future ego trajectory is represented as
$a_0\in\mathbb{R}^{L\times3}$, consisting of waypoints
$(x_\ell,y_\ell,\psi_\ell)$, and is normalized before flow matching.
Given an independently sampled action flow time
$t_a\sim\mathcal{U}(0,1)$ and noise
$\epsilon_a\sim\mathcal{N}(0,I)$, we construct
\begin{equation}
a_{t_a}=(1-t_a)a_0+t_a\epsilon_a.
\label{eq:act-noise}
\end{equation}
The noised trajectory and structured driving context are encoded as
\begin{equation}
h_{\mathrm{act}}=E_{\mathrm{act}}(a_{t_a}),
\qquad
h_{\mathrm{ctx}}=E_{\mathrm{ctx}}([s_{\leq0};g]).
\label{eq:act-enc}
\end{equation}
Conditioned on the driving context and video features, the Action DiT predicts
\begin{equation}
v_\phi^a(a_{t_a},t_a,c^a,\mathcal{F})
=
\mathrm{DiT}_{\mathrm{act}}
\left(
[h_{\mathrm{act}};t_a]
\,\middle|\,
h_{\mathrm{ctx}},\mathcal{F}
\right),
\end{equation}
where $c^a=\{s_{\leq0},g\}$ is represented by $h_{\mathrm{ctx}}$.
The corresponding flow-matching objective~\cite{lipman2022flow} is
\begin{equation}
\mathcal{L}_{\mathrm{FM}}
=
\mathbb{E}_{a_0,t_a,\epsilon_a}
\left[
\left\|
v_\phi^a(a_{t_a},t_a,c^a,\mathcal{F})
-(\epsilon_a-a_0)
\right\|_2^2
\right].
\label{eq:fm}
\end{equation}
The standard chained WAM is trained sequentially with
\begin{equation}
\underbrace{\mathcal{L}_{\mathrm{Gen}}}_{\text{video training}}
\quad\longrightarrow\quad
\underbrace{\mathcal{L}_{\mathrm{FM}}}_{\text{action training}}.
\label{eq:std}
\end{equation}

During planning inference, the initial video latent is sampled from
$\mathcal{N}(0,I)$, and a single Video DiT forward pass at the first discrete denoising step extracts $\mathcal{F}$. The Action DiT then generates the trajectory without decoding the future video, retaining the computational efficiency of latent-space planning. During training, the video and action flow times $t_v$ and $t_a$ are sampled independently. In Stage~2, the Video DiT parameters are frozen. In Stage~3, $\mathcal{F}$ is recomputed online without gradient detachment, allowing planning losses to propagate through the Action DiT conditioning pathway into the Video DiT. In the following, $v_i$, $v_f$, $v_w$, and $\gamma$ denote the intermediate velocity, final velocity, self-guided velocity, and guidance scale, respectively, as defined in Eq.~\eqref{eq:selfguide}.

\subsection{Explicitly Shaping the World-to-Action Representation Pathway}
\label{sec:reworld_overview}

A standard chained WAM is optimized using output-level video-generation and trajectory-prediction objectives. Although these objectives provide the supervision required for the two output tasks, they leave the intermediate world-to-action pathway largely unconstrained. In particular, they do not explicitly specify what future information should be exposed by planner-facing Video DiT states, what video information should be retained by Action DiT states after cross-attention, or how the resulting action space should reflect closed-loop behavior quality.

To the best of our knowledge, ReWorld is the first representation learning framework specifically designed for autonomous-driving WAMs. It addresses these three limitations through a progressive curriculum for representation formation, cross-modal transfer, and decision-oriented shaping. First, ReWorld directly supervises intermediate Video DiT states using the future-video flow target, encouraging future-predictive information to emerge before the final prediction head. Second, it freezes the video branch and uses the attended video readout computed in each forward pass as a stop-gradient grounding target for the Action DiT. Third, it introduces a decision-oriented trajectory objective that repels predictions from geometrically close but low-scoring alternatives. The first two stages directly constrain intermediate features, whereas the third shapes the shared representation pathway through hard-negative trajectory gradients.

These objectives are applied sequentially because they play distinct optimization roles. Stage~1 first establishes future-predictive structure in the video representation. Stage~2 then grounds the action representation in a stable video feature space. Finally, Stage~3 allows planning-oriented gradients to jointly adapt the Action DiT and the planner-facing Video DiT states. Accordingly, we apply $\mathcal{L}_{\mathrm{align}}$ and $\mathcal{L}_{\mathrm{RDE}}$ in separate stages rather than combining them into a single objective. Stage~2 requires stop-gradient readouts derived from frozen video features to establish a stable grounding target, whereas Stage~3 intentionally allows decision-oriented gradients to reshape the planner-facing video states.

\subsection{Future-Predictive World Representations}

Video generators acquire powerful world priors because predicting future observations requires internalizing how scenes evolve~\cite{brooks2024video,bruce2024genie,agarwal2025cosmos}. Under standard diffusion training, however, this predictive structure is explicitly enforced only at the final output. Intermediate layers are free to organize information according to the final generation objective, without being directly required to expose future-predictive structure. This creates a mismatch in a chained WAM because the planner consumes precisely these intermediate Video DiT states. Inspired by~\cite{zhou2026guiding}, we attach auxiliary prediction heads to selected intermediate Video DiT layers.

Let $h_{t_v}^{(l)}$ denote the hidden feature of the $l$-th Video DiT block at video flow time $t_v$. For every supervised layer $l\in\mathcal{S}$, a lightweight head $q_l(\cdot)$ predicts the same velocity target as the final generation head:
\begin{equation}
\hat{v}_{t_v}^{(l)}
=
q_l\!\left(h_{t_v}^{(l)}\right).
\label{eq:midhead}
\end{equation}
Unless otherwise specified, we use $\mathcal{S}=\{8\}$. The intermediate supervision objective is
\begin{equation}
\mathcal{L}_{\mathrm{Mid}}
=
\frac{1}{|\mathcal{S}|}
\sum_{l\in\mathcal{S}}
\mathbb{E}_{z_0,t_v,\epsilon_z}
\left[
\left\|
\hat{v}_{t_v}^{(l)}-v_{t_v}^{*}
\right\|_2^2
\right],
\qquad
v_{t_v}^{*}=\epsilon_z-z_0,
\label{eq:mid}
\end{equation}
and the Stage~1 objective becomes
\begin{equation}
\mathcal{L}_{\mathrm{Video}}
=
\mathcal{L}_{\mathrm{Gen}}
+
\lambda_{\mathrm{Mid}}\mathcal{L}_{\mathrm{Mid}}.
\label{eq:stage1}
\end{equation}

Intermediate supervision is the representation-learning objective, whereas self-guidance is an inference-time mechanism enabled by the resulting cross-layer prediction hierarchy. During training, $\mathcal{L}_{\mathrm{Mid}}$ moves the future-video constraint from the final output into the process of intermediate representation formation. Empirically, this supervision reaches a comparable validation level using approximately half the optimization steps required by vanilla flow matching, corresponding to an approximately $2\times$ convergence acceleration (Fig.~\ref{fig:infer}(b)).

Intermediate supervision also induces a systematic discrepancy between the intermediate and final velocity predictions (Fig.~\ref{fig:infer}(a)). At inference, we use this cross-layer discrepancy as a correction direction. Let $v_i$ and $v_f$ denote the velocities predicted by the supervised intermediate head and final head, respectively. We construct the self-guided velocity as
\begin{equation}
v_w
=
v_i+\gamma\left(v_f-v_i\right),
\label{eq:selfguide}
\end{equation}
where $\gamma$ is the guidance scale. The scheduler uses $v_w$ in place of $v_f$ to advance the denoising trajectory. This correction is applied only during sampling and does not alter the training objective. The Action DiT continues to read $\mathcal{F}$ from the first discrete video denoising step of the same Video DiT. Therefore, self-guidance improves video sampling without changing the planner-facing latent interface. The reported FVD of 61.9 reflects the combined effect of intermediate supervision and self-guided sampling.

\subsection{World-Grounded Action Representations}

The chained interface allows action tokens to retrieve video information through cross-attention, but access alone does not guarantee retention. Under standard trajectory supervision, the Action DiT may use attended video information transiently without preserving it in the post-attention states that support subsequent trajectory generation. We therefore introduce an explicit alignment objective that encourages Action DiT states to retain information consistent with their attended video readouts.

At the $k$-th cross-attention layer of the Action DiT, let $a_i^{(k)}$ denote the post-cross-attention hidden state of action token $i$. Let $\alpha_{ij}^{(k)}$ denote its attention weight on video token $j$, and let $v_j^{(k)}$ denote the corresponding value vector. For notational simplicity, we write the output-projected multi-head cross-attention readout as
\begin{equation}
r_i^{(k)}
=
\sum_j\alpha_{ij}^{(k)}v_j^{(k)}.
\label{eq:readout}
\end{equation}
We encourage the post-attention action state to remain consistent with this retrieved video information in representation space. Empirically, we apply the loss to a deep Action DiT cross-attention layer, $\mathcal{K}=\{12\}$, where video information has already been aggregated into action states that directly support trajectory flow matching:
\begin{equation}
\mathcal{L}_{\mathrm{align}}
=
\frac{1}{|\mathcal{K}|N_a}
\sum_{k\in\mathcal{K}}
\sum_{i=1}^{N_a}
\left[
1-\cos\!\left(
a_i^{(k)},
\operatorname{sg}(r_i^{(k)})
\right)
\right].
\label{eq:align}
\end{equation}
Here, $N_a$ is the number of action tokens, and cosine similarity measures directional agreement in representation space. The stop-gradient operator prevents the readout target from adapting merely to reduce the alignment loss. Gradients continue to flow through $a_i^{(k)}$, including its Action DiT cross-attention pathway, so the planner is trained to preserve the attended information rather than allowing the grounding target to follow the planner state.

In Stage~2, the Video DiT is frozen and only the Action DiT is updated:
\begin{equation}
\mathcal{L}_{\mathrm{act}}^{(2)}
=
\mathcal{L}_{\mathrm{FM}}
+
\lambda_{\mathrm{align}}\mathcal{L}_{\mathrm{align}}.
\label{eq:stage2}
\end{equation}
Freezing the video branch establishes a stable world-representation space. The attended readout is recomputed in every forward pass, and its attention weights still depend on the Action DiT, but the stop-gradient operation makes the resulting readout a fixed grounding target within each optimization step.

\subsection{Behavior-Aware Action Representations}

World-grounded action states encode scene content and future dynamics, but they are not explicitly organized according to closed-loop behavior quality. A trajectory may be geometrically close to the expert while still causing collisions, violating drivable-area constraints, or reducing progress and comfort. Standard imitation learning provides little contrastive supervision among such nearby alternatives. Alignment with video readouts cannot provide this signal either because the video prior is not optimized using closed-loop quality scores. Inspired by~\cite{wang2026beyond}, we introduce hard-negative trajectory supervision to make the world-to-action pathway sensitive to low-scoring closed-loop behaviors.

\noindent\textbf{Hard-negative construction.}
For each training scene, we construct an offline pool of $N$ candidate trajectories
$\{\tau^{(n)}\}_{n=1}^{N}$, where
$\tau^{(n)}\in\mathbb{R}^{L\times3}$, and evaluate them using the NAVSIM PDM simulator~\cite{dauner2024navsim}. We use the overall PDM score as the closed-loop quality function $s(\cdot)$, with higher values indicating better aggregate behavior. For an expert trajectory $\tau^{\mathrm{exp}}$, the hard negative $\tau^{\mathrm{neg}}$ is defined as the closest low-scoring candidate in normalized trajectory space:
\begin{gather}
\mathcal{I}_{\mathrm{low}}
=
\left\{
n\mid s(\tau^{(n)})<\delta
\right\},
\qquad
\delta=0.6,
\label{eq:low}\\[2pt]
n^\star
=
\arg\min_{n\in\mathcal{I}_{\mathrm{low}}}
\frac{1}{L}
\sum_{\ell=1}^{L}
\left\|
\tau_\ell^{(n)}-\tau_\ell^{\mathrm{exp}}
\right\|_2^2,
\label{eq:nstar}\\[2pt]
\tau^{\mathrm{neg}}
=
\tau^{(n^\star)}.
\label{eq:tauneg}
\end{gather}
The nearest-neighbor distance is computed after applying the same per-channel normalization used for trajectory flow matching. Scenes without a low-scoring candidate are excluded from this objective. Selecting the nearest low-scoring trajectory, rather than a random low-scoring candidate, matters because random negatives are often distinguishable by geometry alone. The selected hard negative instead occupies the local neighborhood of the expert while scoring poorly in closed loop, exposing precisely the behavioral ambiguity that standard imitation does not resolve.

\noindent\textbf{Repulsive distance loss.}
Because the planner parameterizes trajectories through a velocity field, we recover an instantaneous clean trajectory estimate from the same forward pass and the same sampled $t_a$ used by $\mathcal{L}_{\mathrm{FM}}$:
\begin{equation}
\hat{a}_0
=
a_{t_a}
-
t_a v_\phi^a(a_{t_a},t_a,c^a,\mathcal{F}),
\qquad
\hat{\tau}
=
\operatorname{Denorm}(\hat{a}_0),
\label{eq:cleanest}
\end{equation}
where $\operatorname{Denorm}(\cdot)$ is the differentiable affine inverse of the normalization used for trajectory flow matching. To emphasize relative motion rather than absolute position, we use a delta representation. Each waypoint is mapped to
\begin{equation}
\Delta(\tau)_\ell
=
\left[
\widetilde{\Delta x}_\ell,
\widetilde{\Delta y}_\ell,
\sin\psi_\ell,
\cos\psi_\ell
\right]
\in\mathbb{R}^{4},
\label{eq:delta}
\end{equation}
where $\widetilde{\Delta x}_\ell$ and $\widetilde{\Delta y}_\ell$ are normalized position increments. The sine--cosine parameterization avoids the discontinuity of angular coordinates. We then define the repulsive distance objective as
\begin{equation}
\mathcal{L}_{\mathrm{RDE}}
=
-\frac{1}{|\mathcal{V}|}
\sum_{b\in\mathcal{V}}
\frac{1}{L}
\sum_{\ell=1}^{L}
\frac{1}{4}
\sum_{d=1}^{4}
\left|
\Delta(\hat{\tau}_b)_\ell^{(d)}
-
\Delta(\tau_b^{\mathrm{neg}})_\ell^{(d)}
\right|,
\label{eq:rde}
\end{equation}
where $\mathcal{V}$ denotes the set of training scenes with valid hard negatives, and $d$ indexes the four channels in Eq.~\eqref{eq:delta}. If no valid hard negative exists in a minibatch, i.e., $\mathcal{V}=\varnothing$, we set $\mathcal{L}_{\mathrm{RDE}}=0$.

The negative sign encourages the predicted trajectory to move away from the selected low-scoring neighbor in delta space, while $\mathcal{L}_{\mathrm{FM}}$ continues to anchor it to the expert trajectory. Although the repulsive objective is not lower-bounded in isolation, it is used only as a weak regularizer alongside the quadratic flow-matching objective. As shown by the weight analysis in Sec.~\ref{sec:reworld_ablations}, excessive repulsion competes with imitation and degrades closed-loop performance. Gradients from $\mathcal{L}_{\mathrm{RDE}}$ propagate through the shared forward graph into the Action DiT. When the Video DiT is unfrozen in Stage~3, $\mathcal{F}$ is recomputed online without detachment, allowing these behavior-oriented gradients to adapt the video features used for planning.

In Stage~3, the Video DiT and Action DiT are jointly fine-tuned:
\begin{equation}
\mathcal{L}_{\mathrm{act}}^{(3)}
=
\mathcal{L}_{\mathrm{FM}}
+
\lambda_{\mathrm{RDE}}\mathcal{L}_{\mathrm{RDE}}.
\label{eq:stage3}
\end{equation}
This stage follows Stage~2 and does not retain $\mathcal{L}_{\mathrm{align}}$, because the planner-facing video representation is intentionally allowed to adapt under behavior-oriented supervision.

\subsection{Progressive Representation Curriculum}
\label{sec:training_protocol}

ReWorld applies a three-stage representation curriculum. Stage~1 trains the Video DiT with future-predictive intermediate supervision using $\mathcal{L}_{\mathrm{Video}}$ (Eq.~\eqref{eq:stage1}). Stage~2 freezes the Video DiT and grounds the Action DiT in attended video information using $\mathcal{L}_{\mathrm{act}}^{(2)}$ (Eq.~\eqref{eq:stage2}). Stage~3 jointly fine-tunes both branches with hard-negative repulsion using $\mathcal{L}_{\mathrm{act}}^{(3)}$ (Eq.~\eqref{eq:stage3}). This ordering first forms a future-predictive world representation, then establishes stable world-to-action transfer, and finally shapes the shared pathway according to closed-loop behavior quality. Initialization, step counts, batch sizes, and hyperparameters are provided in Sec.~\ref{sec:experiment}. Hard negatives are mined offline using training scenes only. During video sampling, ReWorld applies self-guidance with $\gamma=1.4$, while the Action DiT continues to read $\mathcal{F}$ from the first discrete video denoising step during planning.

\section{Experiments}
\label{sec:experiment}

\subsection{Experimental Setup}
\label{sec:setup}

\noindent\textbf{Tasks, datasets, and metrics.}
Following established autonomous-driving WAM protocols~\cite{zhang2025epona,li2025drivevla,zhao2026forecasting}, we use nuPlan~\cite{caesar2021nuplan} and nuScenes~\cite{caesar2020nuscenes} for video training and NAVSIM~\cite{dauner2024navsim} for trajectory learning and evaluation. nuScenes contains 1{,}000 urban driving sequences collected in Boston and Singapore with synchronized camera and LiDAR observations, including 850 development sequences and 150 held-out sequences. nuPlan provides approximately 1{,}200 hours of human-driving data collected across four metropolitan areas. We sample camera streams at $8\,\mathrm{Hz}$ for video training and use $2\,\mathrm{Hz}$ observations from NAVSIM for trajectory supervision.

NAVSIM is a data-driven, non-reactive closed-loop planning benchmark built on OpenScene~\cite{contributors2023openscene}, with approximately 103k scenes in \textit{Navtrain} and 12k scenes in \textit{Navtest}. We evaluate video generation on nuScenes using Fr\'echet Inception Distance (FID)~\cite{heusel2017gans} and Fr\'echet Video Distance (FVD)~\cite{unterthiner2018towards}. FID measures frame-level visual fidelity, while FVD evaluates video quality with an emphasis on temporal coherence. Temporal ablations therefore focus on FVD. Closed-loop planning is evaluated on NAVSIM v1~\cite{dauner2024navsim} using no-at-fault collision (NC), drivable-area compliance (DAC), time-to-collision (TTC), comfort (Comf.), ego progress (EP), and the Predictive Driver Model Score (PDMS):
\begin{equation}
\mathrm{PDMS}
=
\mathrm{NC}\times\mathrm{DAC}\times
\frac{
5\cdot\mathrm{EP}
+
5\cdot\mathrm{TTC}
+
2\cdot\mathrm{Comf.}
}{12}.
\end{equation}
We further perform frozen linear probing on UCF-101 action recognition to assess the quality and transferability of the spatiotemporal representations learned by the Video DiT.

\noindent\textbf{Backbone and sequential pretraining.}
ReWorld is instantiated on the chained DriveLaW architecture~\cite{xia2026drivelaw}, comprising a 2B Video DiT~\cite{peebles2023scalable} initialized from LTX-Video~\cite{hacohen2024ltx} and a 133M Action DiT. The standard chained WAM is obtained using the sequential objectives in Eq.~\eqref{eq:std}. The Video DiT is first optimized with $\mathcal{L}_{\mathrm{Gen}}$ through a progressive video curriculum consisting of long low-resolution clips ($740\times352\times121$), followed by short high-resolution clips ($1280\times704\times25$). The Action DiT is subsequently conditioned on the pretrained video latents and optimized with $\mathcal{L}_{\mathrm{FM}}$ for trajectory generation.

Throughout the experiments, \textbf{DriveLaW} denotes this standard chained WAM trained with output-level supervision, while \textbf{ReWorld} augments the same architecture and latent interface with the proposed representation curriculum.

\noindent\textbf{Representation curriculum.}
Stage~1 initializes the Video DiT from LTX-Video~\cite{hacohen2024ltx} and optimizes it with
$\mathcal{L}_{\mathrm{Gen}}
+\lambda_{\mathrm{Mid}}\mathcal{L}_{\mathrm{Mid}}$
for 20k steps using a global batch size of 64. We use AdamW with a learning rate of $1\times10^{-5}$ and a weight decay of $5\times10^{-2}$. The video flow time is sampled token-wise from $t_v\in[0,1]$, and the default supervised block is $\mathcal{S}=\{8\}$.

Stage~2 initializes from the DriveLaW checkpoint, freezes the Video DiT, and optimizes the Action DiT with
$\mathcal{L}_{\mathrm{FM}}
+\lambda_{\mathrm{align}}\mathcal{L}_{\mathrm{align}}$
for 6k steps using a batch size of 128. We set
$\lambda_{\mathrm{align}}=0.05$
and apply the alignment objective at the 12th cross-attention layer.

Stage~3 jointly fine-tunes the Video DiT and Action DiT with
$\mathcal{L}_{\mathrm{FM}}
+\lambda_{\mathrm{RDE}}\mathcal{L}_{\mathrm{RDE}}$
for 10k steps using a batch size of 160, with
$\lambda_{\mathrm{RDE}}=0.04$.
Video features are recomputed online, allowing decision-oriented gradients to shape the planner-facing world representations.

Hard negatives are mined offline from the training set following BeyondDrive~\cite{wang2026beyond}. For each scene, a flow-matching trajectory generator produces 64 candidate trajectories using classifier-free guidance and noise-scale adjustment to increase diversity. Each candidate is evaluated by the NAVSIM PDM simulator. Candidates with scores below $\delta=0.6$ form the low-scoring set, from which we select the trajectory closest to the expert under the same per-channel normalization used for trajectory flow matching. Video generation uses 30 sampling steps with self-guidance scale $\gamma=1.4$, and trajectory generation uses five flow-matching steps.

\noindent\textbf{Frozen linear probing.}
We evaluate frozen Video DiT representations on UCF-101 split~1. All checkpoints follow the same protocol, using 33-frame clips at $224\times224$, the same video flow timestep, feature normalization, and classifier schedule. Given VAE-encoded video latents, we extract the representation from the final block of the 28-layer Video DiT,
\begin{equation}
h^{(28)}\in\mathbb{R}^{B\times N\times C_h},
\end{equation}
apply global mean pooling over the token dimension, and train a linear classifier on the resulting $\mathbb{R}^{B\times C_h}$ features. We compare LTX-Video, DriveLaW, ReWorld after Stage~1, and ReWorld after the full curriculum.

\noindent\textbf{Unified representation-learning protocol.}
For controlled comparison with diffusion representation-learning methods, we train all approaches from scratch on LTX-Video using nuPlan and nuScenes for 120k steps. The unified setting uses a batch size of 32, $224\times224\times25$ video clips, and no text encoder. We report FVD on the nuScenes test set and normalized per-step Video DiT training cost relative to vanilla flow matching.

\subsection{Main Results}
\label{sec:main_results}

We first evaluate the two functional endpoints of a WAM. Video generation measures the quality of future-world modeling, while closed-loop planning evaluates how effectively the learned world representation supports action generation.

\noindent\textbf{Video generation.}
Tab.~\ref{tab:video_generation} reports video-generation results on the nuScenes validation set. DriveLaW establishes a strong baseline with 4.6 FID and 81.3 FVD, outperforming previous single-view generators including DriveWorld~\cite{min2024driveworld}, Vista~\cite{gao2024vista}, and Epona~\cite{zhang2025epona}.

With standard sampling ($\gamma=1.0$), future-predictive intermediate supervision improves FVD from 81.3 to 78.9, showing that direct supervision of intermediate Video DiT states strengthens temporal generation. The resulting intermediate-to-final prediction hierarchy further enables self-guided sampling, reducing FVD to 61.9, a relative improvement of 23.9\% over DriveLaW. ReWorld also improves FID from 4.6 to 4.4.

The FVD reduction indicates stronger modeling of scene dynamics and cross-frame consistency, while the FID result shows that ReWorld preserves the frame-level fidelity of the underlying video generator. Together, these results support future-predictive intermediate supervision as a mechanism for improving both the sampling process and the generated future.

\begin{table}[t]
\caption{\textbf{Video generation on the nuScenes validation set.}
DriveLaW denotes the standard chained WAM trained with the sequential objectives in Eq.~\eqref{eq:std}. ReWorld uses future-predictive intermediate supervision and self-guided sampling with $\gamma=1.4$.}
\label{tab:video_generation}
\centering
\setlength{\tabcolsep}{14pt}
\renewcommand{\arraystretch}{1.15}
\begin{tabular}{l | c c}
\toprule
\textbf{Method} & \textbf{FID$\downarrow$} & \textbf{FVD$\downarrow$} \\
\midrule
DriveGAN~\cite{kim2021drivegan} & 73.4 & 502.3 \\
DriveDreamer~\cite{wang2024drivedreamer} & 52.6 & 452.0 \\
DrivingGPT~\cite{chen2025drivinggpt} & 12.8 & 142.6 \\
DriveWorld~\cite{min2024driveworld} & 7.4 & 90.9 \\
Vista~\cite{gao2024vista} & 6.9 & 89.4 \\
Epona~\cite{zhang2025epona} & 7.5 & 82.8 \\
DriveLaW~\cite{xia2026drivelaw} & 4.6 & 81.3 \\
\textbf{ReWorld (Ours)} & \textbf{4.4} & \textbf{61.9} \\
\bottomrule
\end{tabular}
\end{table}

\noindent\textbf{Closed-loop planning.}
Tab.~\ref{tab:planning_navsim} reports closed-loop planning results on NAVSIM \textit{Navtest}. ReWorld improves PDMS from 89.1 to 90.4 on the same chained WAM architecture, achieving the best overall performance among the compared methods. The gains are particularly evident in safety- and compliance-related metrics: DAC increases from 97.1 to 98.2, TTC from 96.7 to 97.7, and NC from 99.0 to 99.1.

ReWorld outperforms strong traditional end-to-end planners, including the camera--LiDAR DiffusionDrive~\cite{liao2025diffusiondrive}, as well as world-model-based approaches such as Epona~\cite{zhang2025epona}, DriveVLA-W0~\cite{li2025drivevla}, PWM~\cite{zhao2026forecasting}, and WorldDrive~\cite{gui2026bridging}. It achieves the highest PDMS, NC, DAC, and TTC among the compared world-model methods. These results indicate that explicitly learning world-to-action transfer and behavior-sensitive action representations converts predictive world knowledge into safer closed-loop decisions.

\begin{table*}[t]
\caption{\textbf{Closed-loop planning performance on NAVSIM \textit{Navtest}.} 
Methods are grouped into traditional end-to-end planners and world-model methods. 
$^{\dagger}$ denotes training with the same flow-matching objective. 
ReWorld applies the proposed representation curriculum to the chained DriveLaW architecture.}
\label{tab:planning_navsim}
\centering
\setlength{\tabcolsep}{10pt}
\begin{tabular}{@{}l|c|cc|ccccc|c@{}}
\toprule
\textbf{Method} & \textbf{Ref} & \textbf{Image} & \textbf{Lidar}
& \textbf{NC$\uparrow$} & \textbf{DAC$\uparrow$}
& \textbf{TTC$\uparrow$} & \textbf{Comf.$\uparrow$}
& \textbf{EP$\uparrow$} & \textbf{PDMS$\uparrow$} \\
\midrule
\multicolumn{10}{@{}l}{\textit{Traditional End-to-End Methods}} \\
VADv2-$\mathcal{V}_{\text{8192}}$~\cite{chen2024vadv2}
& arXiv'24 & \checkmark & & 97.2 & 89.1 & 91.6 & \textbf{100} & 76.0 & 80.9 \\
UniAD~\cite{hu2023planning}
& CVPR'23 & \checkmark & & 97.8 & 91.9 & 92.9 & \textbf{100} & 78.8 & 83.4 \\
TransFuser~\cite{chitta2022transfuser}
& TPAMI'23 & \checkmark & \checkmark & 97.7 & 92.8 & 92.8 & \textbf{100} & 79.2 & 84.0 \\
PARA-Drive~\cite{weng2024drive}
& CVPR'24 & \checkmark & & 97.9 & 92.4 & 93.0 & 99.8 & 79.3 & 84.0 \\
ReCogDrive-IL~\cite{li2025recogdrive}
& ICLR'26 & \checkmark & & 98.1 & 94.7 & 94.2 & \textbf{100} & 80.9 & 86.5 \\
DiffusionDrive~\cite{liao2025diffusiondrive}
& CVPR'25 & \checkmark & \checkmark & 98.2 & 96.2 & 94.7 & \textbf{100} & 82.2 & 88.1 \\
\midrule
\multicolumn{10}{@{}l}{\textit{World Model Methods}} \\
DrivingGPT~\cite{chen2025drivinggpt}
& ICCV'25 & \checkmark & & 98.9 & 90.7 & 94.9 & 95.6 & 79.7 & 82.4 \\
LAW~\cite{li2024enhancing}
& ICLR'25 & \checkmark & & 96.4 & 95.4 & 88.7 & 99.9 & 81.7 & 84.6 \\
Epona~\cite{zhang2025epona}
& ICCV'25 & \checkmark & & 97.9 & 95.1 & 93.8 & 99.9 & 80.4 & 86.2 \\
ReSim~\cite{yang2025resim}
& NeurIPS'25 & \checkmark & & -- & -- & -- & -- & -- & 86.6 \\
WoTE~\cite{li2025end}
& ICCV'25 & \checkmark & \checkmark & 98.5 & 96.8 & 94.9 & 99.9 & 81.9 & 88.3 \\
DriveVLA-W0$^{\dagger}$~\cite{li2025drivevla}
& ICLR'26 & \checkmark & & 98.4 & 95.3 & 95.2 & \textbf{100} & 80.9 & 87.2 \\
PWM~\cite{zhao2026forecasting}
& NeurIPS'25 & \checkmark & & 98.6 & 95.9 & 95.4 & \textbf{100} & 81.8 & 88.1 \\
WorldDrive~\cite{gui2026bridging}
& arXiv'26 & \checkmark & & 98.4 & 96.8 & 95.2 & \textbf{100} & \textbf{83.3} & 89.0 \\
DriveLaW~\cite{xia2026drivelaw}
& CVPR'26 & \checkmark & & 99.0 & 97.1 & 96.7 & \textbf{100} & 81.3 & 89.1 \\
\textbf{ReWorld (Ours)}
& - & \checkmark & & \textbf{99.1} & \textbf{98.2}
& \textbf{97.7} & 99.8 & 82.0 & \textbf{90.4} \\
\bottomrule
\end{tabular}%
\end{table*}

\begin{figure*}[t]
  \centering
  \includegraphics[width=\textwidth]{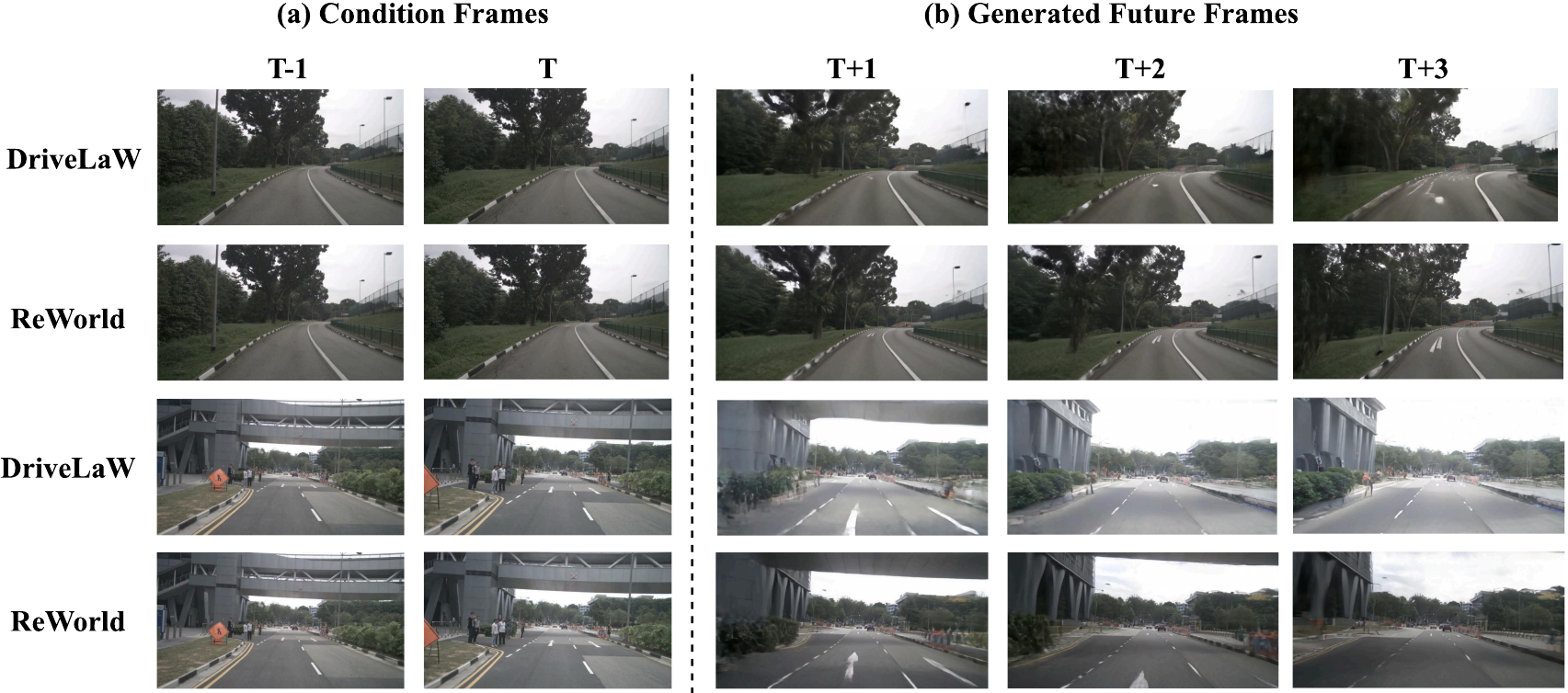}
  \caption{\textbf{Qualitative video-generation comparison with DriveLaW~\cite{xia2026drivelaw}.}
  Conditioning uses $1\,\mathrm{s}$ history ($8$ frames at $8\,\mathrm{Hz}$); columns $T{-}1$ and $T$ show only its first and last frames for visualization.
  Generated future frames ($3\,\mathrm{s}$, $24$ frames) lie to the right of the dashed line (columns $T{+}1$ to $T{+}3$).
  Each pair of consecutive rows depicts one scene for DriveLaW and ReWorld, respectively.
  ReWorld better preserves lane markings, roadside geometry, distant objects, and temporal consistency.}
  \label{fig:qualitative}
\end{figure*}

Fig.~\ref{fig:qualitative} provides a qualitative comparison of future video generation between ReWorld and DriveLaW.
Both methods are conditioned on one second of historical observation ($8$ frames at $8\,\mathrm{Hz}$) and synthesize the subsequent three seconds ($24$ frames).
In the figure, the dashed line separates the conditioning interval from the generated future, and each scene is shown as a DriveLaW--ReWorld pair.
For visualization, columns $T{-}1$ and $T$ show only the first and last frames of the $1\,\mathrm{s}$ history, while columns $T{+}1$ to $T{+}3$ depict the predicted $3\,\mathrm{s}$ future.
Relative to DriveLaW, ReWorld yields sharper lane markings, more stable roadside structures, clearer distant agents, and stronger temporal continuity over the predicted horizon, which is consistent with the observed FVD reduction and indicates that future-predictive intermediate supervision improves long-horizon video coherence under challenging dynamics.

\subsection{Representation Analysis}
\label{sec:representation_analysis}

The task-level results show gains at both outputs of the WAM. We next examine the representations underlying these improvements from two perspectives: their transferability to motion recognition and their behavior under a unified generative representation-learning protocol.

\noindent\textbf{Frozen linear probing on UCF-101.}
Tab.~\ref{tab:ucf101_linear_probe} reports frozen linear-probe accuracy on UCF-101. LTX-Video achieves \ucfltx\%, while driving-domain pretraining in DriveLaW improves the accuracy to \ucfdrivelaw\%. Stage~1 future-predictive supervision further increases it to \ucfreworld\%, demonstrating that intermediate flow supervision strengthens the motion-discriminative structure encoded by the Video DiT.

After the full representation curriculum, ReWorld reaches \ucfreworldfull\%, improving by 11.9 points over DriveLaW. The progression from generic video pretraining to driving adaptation and then to explicit representation learning shows that ReWorld yields stronger spatiotemporal representations whose motion structure transfers to action recognition.

\begin{table}[t]
\caption{\textbf{Frozen linear probing on UCF-101 split~1.}
All methods use 33-frame clips at $224\times224$, final-block Video DiT features, global mean pooling, and the same linear-classifier protocol.}
\label{tab:ucf101_linear_probe}
\centering
\setlength{\tabcolsep}{6pt}
\begin{tabular}{l|c}
\toprule
\textbf{Frozen Video DiT representation}
& \textbf{Top-1 Acc. (\%)$\uparrow$} \\
\midrule
LTX-Video~\cite{hacohen2024ltx} & \ucfltx \\
DriveLaW~\cite{xia2026drivelaw} & \ucfdrivelaw \\
ReWorld (Stage~1) & \ucfreworld \\
\textbf{ReWorld} & \textbf{\ucfreworldfull} \\
\bottomrule
\end{tabular}
\end{table}

\noindent\textbf{Comparison with representation-learning methods.}
Tab.~\ref{tab:repr_learning_comparison} compares representation-learning strategies under the unified long-horizon driving-video protocol. ReWorld achieves the best FVD of 270.4, outperforming Vanilla Flow by 33.7 points and the strongest competing self-supervised method, Self-Flow, by 12.9 points.

The comparison highlights the importance of task-aligned representation objectives for generative modeling. External feature spaces primarily encode semantic, geometric, or recognition-oriented structure. Long-horizon video generation, however, requires intermediate states to preserve continuous temporal evolution and flow-consistent dynamics. ReWorld directly applies the generator's native future-flow target to intermediate states, aligning representation learning with the temporal structure of the generative process. This generator-native supervision yields the strongest video-generation performance among the compared methods.

\begin{table}[t]
\caption{\textbf{Representation learning for long-horizon driving-video generation.}
All methods are trained from scratch for 120k steps on nuPlan and nuScenes using $224\times224\times25$ clips without text conditioning.}
\label{tab:repr_learning_comparison}
\centering
\setlength{\tabcolsep}{6pt}
\begin{tabular}{l | c | c}
\toprule
\textbf{Model} & \textbf{Steps} & \textbf{FVD$\downarrow$} \\
\midrule
\multicolumn{3}{@{}l}{\textit{Without external representations}} \\
Vanilla Flow & 120k & 304.1 \\
SRA~\cite{jiang2025no} & 120k & 296.9 \\
SRA2~\cite{wang2026sra} & 120k & 295.2 \\
Self-Flow~\cite{chefer2026self} & 120k & 283.3 \\
\textbf{ReWorld (Ours)} & 120k & \textbf{270.4} \\
\midrule
\multicolumn{3}{@{}l}{\textit{With external representations}} \\
REPA w/ DINOv2~\cite{yu2024representation,oquab2023dinov2}
& 120k & 295.9 \\
REPA w/ VideoMAEv2~\cite{yu2024representation,wang2023videomae}
& 120k & 328.3 \\
REPA w/ DepthAnything3~\cite{yu2024representation,lin2025depth}
& 120k & 319.4 \\
REPA w/ V-JEPA2~\cite{yu2024representation,assran2025v}
& 120k & 331.6 \\
ReDi~\cite{kouzelis2026boosting} & 120k & 421.7 \\
\bottomrule
\end{tabular}
\end{table}

\noindent\textbf{Training efficiency.}
Tab.~\ref{tab:efficiency} compares the normalized per-step cost of video-side representation learning. ReWorld introduces only a lightweight intermediate prediction head and incurs $1.003\times$ training cost. In comparison, SRA and Self-Flow require an additional DiT forward pass to construct self-supervised targets, while external-alignment methods require a separate representation encoder.

ReWorld therefore combines the best FVD in Tab.~\ref{tab:repr_learning_comparison} with only 0.3\% additional per-step Video DiT computation. Its efficiency follows from reusing the future-flow target and intermediate activations already available in the generative model.

\begin{table}[t]
\caption{\textbf{Normalized per-step cost of video-side representation learning.}
Costs are measured at $224\times224\times25$ with batch size 1, using vanilla flow matching as $1.0\times$.}
\label{tab:efficiency}
\centering
\small
\setlength{\tabcolsep}{6pt}
\begin{tabular}{@{}
  w{c}{\dimexpr0.46\columnwidth-2\tabcolsep-0.5\arrayrulewidth\relax}
  |
  w{c}{\dimexpr0.46\columnwidth-2\tabcolsep-0.5\arrayrulewidth\relax}
@{}}
\toprule
\textbf{Method} & \textbf{Normalized Training Cost} \\
\midrule
\multicolumn{2}{c}{\textit{Without external representations}} \\
\midrule
Vanilla Flow & 1.0$\times$ \\
Self-Flow~\cite{chefer2026self} & $\sim$1.4$\times$ \\
SRA~\cite{jiang2025no} & $\sim$1.4$\times$ \\
\textbf{ReWorld} & \textbf{1.003$\times$} \\
\midrule
\multicolumn{2}{c}{\textit{With external representations}} \\
\midrule
REPA~\cite{yu2024representation} & $\sim$1.7$\times$ \\
ReDi~\cite{kouzelis2026boosting} & $\sim$1.6$\times$ \\
\bottomrule
\end{tabular}
\end{table}

\begin{figure*}[t]
  \centering
  \includegraphics[width=1\textwidth]{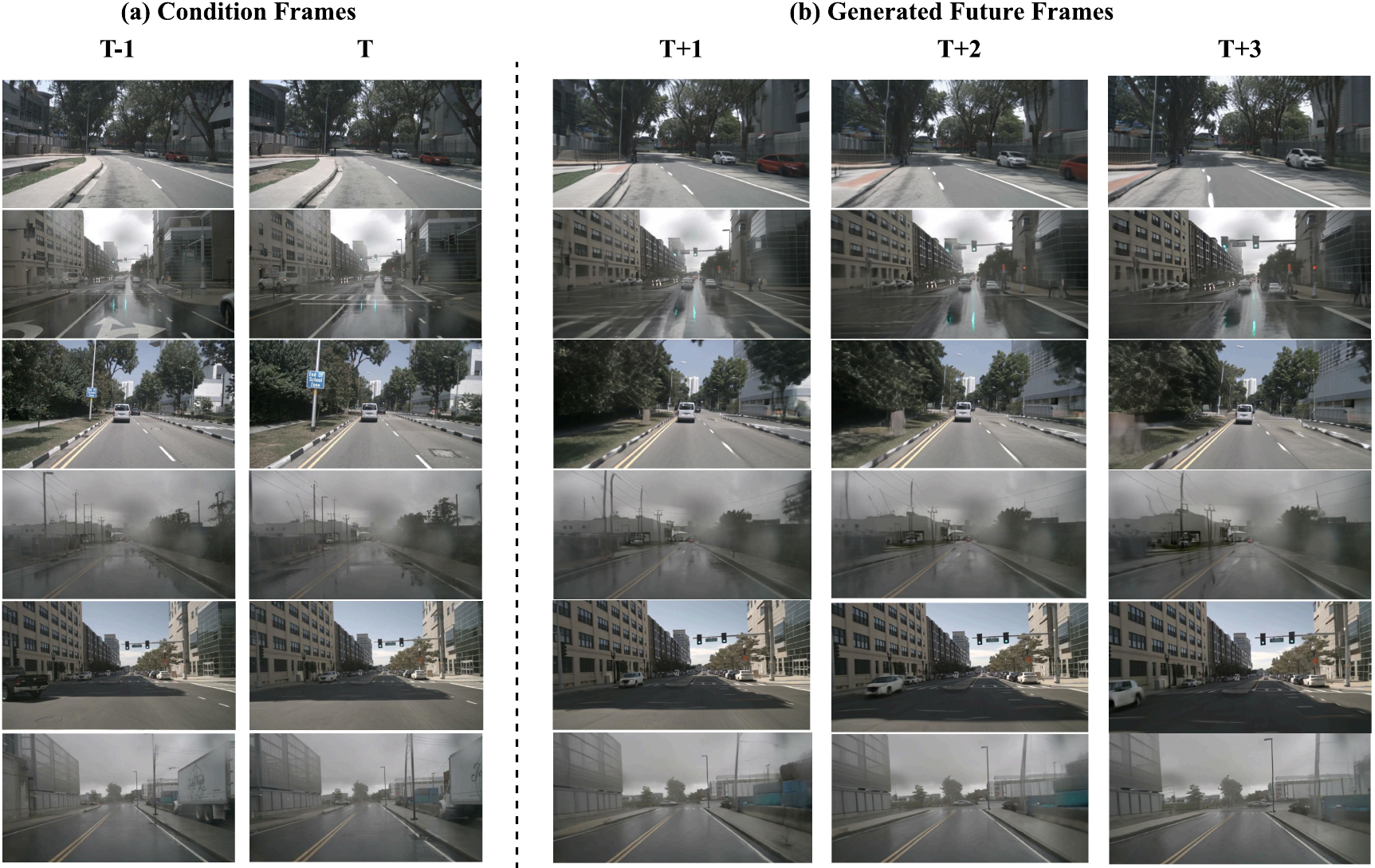}
  \caption{\textbf{Additional video-generation results on nuScenes.}
  Conditioning uses $1\,\mathrm{s}$ history ($8$ frames at $8\,\mathrm{Hz}$); columns $T{-}1$ and $T$ show only its first and last frames for visualization, as in Fig.~\ref{fig:qualitative}.
  Generated futures ($3\,\mathrm{s}$, $24$ frames) appear to the right of the dashed line (columns $T{+}1$ to $T{+}3$).
  Each row corresponds to one driving scene, covering sunny and rainy weather, high-speed travel, intersections, and other urban conditions.
  ReWorld produces temporally coherent futures with stable geometry, clear lane structure, and consistent appearance of surrounding agents.}
  \label{fig:backbone_qualitative}
\end{figure*}

\begin{figure*}[t]
  \centering
  \includegraphics[width=1\textwidth]{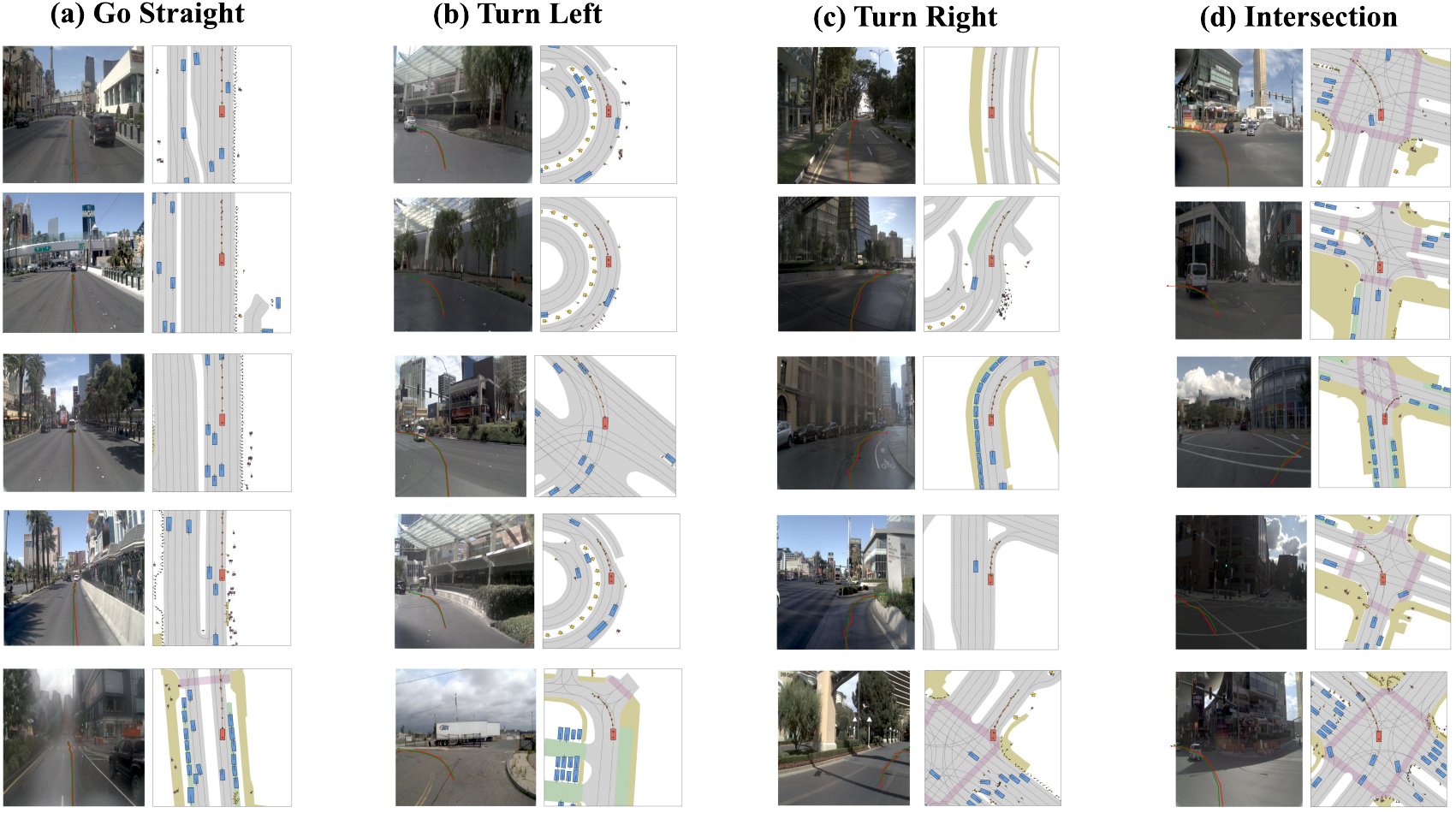}
  \caption{\textbf{Additional planning results on NAVSIM \textit{Navtest}.}
  From left to right: straight, turn left, turn right, and intersection.
  Red curves denote the ego trajectories predicted by our model, and green curves denote the ground-truth expert trajectories.}
  \label{fig:navtest_qualitative}
\end{figure*}

\subsection{Ablation Studies}
\label{sec:reworld_ablations}

We analyze the representation curriculum along its three functional stages: future-predictive world representation, world-to-action transfer, and behavior-aware action shaping.

\noindent\textbf{Future-predictive world representation.}
Intermediate supervision substantially accelerates optimization. As shown in Fig.~\ref{fig:infer}(b), ReWorld reaches the reference validation performance using approximately half the training steps required by vanilla flow matching, corresponding to an approximately $2\times$ convergence acceleration.

Tab.~\ref{tab:ablation_layer_guidance}(a) studies the location of intermediate supervision. Block~8 achieves the best FVD of 61.9, indicating a favorable balance between representation maturity and subsequent refinement. Earlier blocks provide less developed future estimates, while deeper blocks offer less hierarchical separation from the final prediction head.

Tab.~\ref{tab:ablation_layer_guidance}(b) studies the self-guidance scale. Standard sampling with $\gamma=1.0$ yields 78.9 FVD. Increasing $\gamma$ to 1.4 improves FVD to 61.9, showing that the discrepancy between intermediate and final predictions provides an effective refinement direction. The non-monotonic trend further indicates that moderate guidance best exploits the learned prediction hierarchy.

\begin{table}[t]
\caption{\textbf{Intermediate-layer and self-guidance ablations.}
(a)~FVD with different supervised Video DiT blocks.
(b)~FVD under different self-guidance scales using block~8.}
\label{tab:ablation_layer_guidance}
\centering
\setlength{\tabcolsep}{8pt}
\small\textbf{(a) Supervised intermediate block}\\[2pt]
\begin{tabular}{@{}
  w{c}{\dimexpr0.5\ablationshortw-2\tabcolsep-0.5\arrayrulewidth\relax}
  |
  w{c}{\dimexpr0.5\ablationshortw-2\tabcolsep-0.5\arrayrulewidth\relax}
@{}}
\toprule
\textbf{Supervised Block} & \textbf{FVD$\downarrow$} \\
\midrule
2 & 65.5 \\
\textbf{8} & \textbf{61.9} \\
12 & 62.7 \\
16 & 63.0 \\
20 & 64.3 \\
\bottomrule
\end{tabular}
\vspace{6pt}

\small\textbf{(b) Self-guidance scale}\\[2pt]
\begin{tabular}{@{}
  w{c}{\dimexpr0.5\ablationshortw-2\tabcolsep-0.5\arrayrulewidth\relax}
  |
  w{c}{\dimexpr0.5\ablationshortw-2\tabcolsep-0.5\arrayrulewidth\relax}
@{}}
\toprule
\textbf{$\gamma$} & \textbf{FVD$\downarrow$} \\
\midrule
1.0 & 78.9 \\
1.2 & 72.0 \\
\textbf{1.4} & \textbf{61.9} \\
1.6 & 69.7 \\
1.8 & 68.2 \\
\bottomrule
\end{tabular}
\end{table}

\noindent\textbf{World-grounded action representations.}
With the Video DiT frozen, $\mathcal{L}_{\mathrm{align}}$ improves PDMS from 89.1 to 89.5, as shown in Tab.~\ref{tab:ablation_planning_components}. The improvement isolates action-side representation learning and suggests that preserving the attended video readout strengthens world-to-action knowledge transfer.
Tab.~\ref{tab:ablation_align_rde_weight}(a) further studies the alignment weight $\lambda_{\mathrm{align}}$ in Stage~2 only, without Stage~3.
The best closed-loop performance under this Stage~2 setting is obtained at $\lambda_{\mathrm{align}}=0.05$ (PDMS~$89.5$).
Weaker weights provide insufficient grounding of action states to the attended video readout, whereas stronger weights over-constrain the Action DiT and interfere with trajectory flow matching.

\noindent\textbf{Behavior-aware action shaping.}
Applying $\mathcal{L}_{\mathrm{RDE}}$ improves PDMS from 89.1 to 89.8, while progressively combining world grounding and behavior-aware shaping reaches 90.4. The gains confirm their complementary roles: alignment establishes a world-grounded action representation, and RDE further separates locally similar trajectories according to closed-loop behavior quality.

Tab.~\ref{tab:ablation_align_rde_weight}(b) then studies the balance between expert imitation and hard-negative repulsion by training Stage~3 from the best Stage~2 checkpoint above (PDMS~$89.5$ at $\lambda_{\mathrm{align}}=0.05$) and varying only $\lambda_{\mathrm{RDE}}$.
The best performance is obtained at $\lambda_{\mathrm{RDE}}=0.04$.
Smaller weights provide weaker behavioral separation, while larger weights increasingly compete with expert trajectory matching.
This non-monotonic trend shows that RDE is most effective as a local decision-oriented regularizer around the expert trajectory manifold.

\begin{table}[t]
\caption{\textbf{Effects of world grounding and behavior-aware action shaping.}
PDMS is evaluated on NAVSIM \textit{Navtest} using the same chained WAM architecture.}
\label{tab:ablation_planning_components}
\centering
\begin{minipage}{\ablationw}
\centering
\setlength{\tabcolsep}{6pt}
\begin{tabular}{l c c | c}
\toprule
\textbf{Configuration}
& $\mathcal{L}_{\mathrm{align}}$
& $\mathcal{L}_{\mathrm{RDE}}$
& \textbf{PDMS$\uparrow$} \\
\midrule
DriveLaW & & & 89.1 \\
+ Align only & \checkmark & & 89.5 \\
+ RDE only & & \checkmark & 89.8 \\
\textbf{ReWorld}
& \checkmark & \checkmark & \textbf{90.4} \\
\bottomrule
\end{tabular}
\end{minipage}
\end{table}

\begin{table}[t]
\caption{\textbf{Alignment and RDE weight ablations.}
(a)~Stage~2 only (no Stage~3): PDMS under different $\lambda_{\mathrm{align}}$.
(b)~Stage~3 trained from the best Stage~2 checkpoint in~(a) ($\lambda_{\mathrm{align}}=0.05$, PDMS~$89.5$): PDMS under different $\lambda_{\mathrm{RDE}}$.}
\label{tab:ablation_align_rde_weight}
\centering
\setlength{\tabcolsep}{8pt}
\small\textbf{(a) Alignment weight}\\[2pt]
\begin{tabular}{@{}
  w{c}{\dimexpr0.5\ablationshortw-2\tabcolsep-0.5\arrayrulewidth\relax}
  |
  w{c}{\dimexpr0.5\ablationshortw-2\tabcolsep-0.5\arrayrulewidth\relax}
@{}}
\toprule
\textbf{$\lambda_{\mathrm{align}}$} & \textbf{PDMS$\uparrow$} \\
\midrule
0.01 & 88.8 \\
0.03 & 89.2\\
\textbf{0.05} & \textbf{89.5} \\
0.07 & 88.2 \\
0.10 & 87.7 \\
\bottomrule
\end{tabular}
\vspace{6pt}

\small\textbf{(b) RDE weight}\\[2pt]
\begin{tabular}{@{}
  w{c}{\dimexpr0.5\ablationshortw-2\tabcolsep-0.5\arrayrulewidth\relax}
  |
  w{c}{\dimexpr0.5\ablationshortw-2\tabcolsep-0.5\arrayrulewidth\relax}
@{}}
\toprule
\textbf{$\lambda_{\mathrm{RDE}}$} & \textbf{PDMS$\uparrow$} \\
\midrule
0.02 & 89.4 \\
0.03 & 89.6 \\
\textbf{0.04} & \textbf{90.4} \\
0.05 & 89.7 \\
0.10 & 85.5 \\
\bottomrule
\end{tabular}
\end{table}

\subsection{Qualitative and Additional Results}
\label{sec:additional_results}

Fig.~\ref{fig:backbone_qualitative} presents additional generation results of ReWorld on nuScenes under the same conditioning protocol as Fig.~\ref{fig:qualitative}.
Across six representative scenes spanning clear and rainy weather, high-speed driving, intersections, and other common urban settings, the synthesized futures remain temporally coherent, with stable lane geometry, well-preserved roadside structure, and consistent multi-agent appearance over the three-second horizon.

Fig.~\ref{fig:navtest_qualitative} further shows closed-loop planning examples on NAVSIM \textit{Navtest} for straight driving, left turn, right turn, and intersection scenarios.
Red curves denote trajectories predicted by our model and green curves denote ground-truth expert paths; the predicted plans remain smooth and consistent with the surrounding scene layout.

\section{Conclusion}
\label{sec:conclusion}

We presented ReWorld, the first representation learning framework for autonomous-driving World Action Models, which targets the under-constrained intermediate representations along the latent world-to-action pathway. ReWorld explicitly shapes the latent world-to-action pathway through a progressive curriculum: intermediate supervision accelerates convergence, cross-modal alignment preserves the attended world information in action states, and hard-negative shaping increases sensitivity to nearby unsafe behaviors. Requiring no external encoders, the method adds only 0.3\% per-step Video DiT training cost. Across generation, closed-loop planning, and out-of-distribution probing, the results support explicit representation optimization as a central factor in translating world knowledge into planning capability. Future work will explore extensions to longer temporal horizons and multi-modal scenarios.

\section*{Acknowledgments}
This work was in part supported by the National Natural Science Foundation of China (NSFC U25B2067).


\bibliographystyle{IEEEtran}
\bibliography{iclr2026_conference}

\newpage

\begin{IEEEbiography}[{\biosphoto{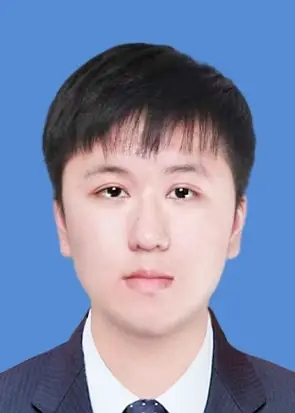}}]{Tianze Xia}
received the B.S.\ degree in Computer Science and Technology from the Huazhong University of Science and Technology, Wuhan, China, in 2026.
He is currently pursuing the M.S.\ degree in electronics and information engineering at the Huazhong University of Science and Technology.
His research interests include autonomous driving, world models, and video generation.
\end{IEEEbiography}
\begin{IEEEbiography}[{\biosphoto{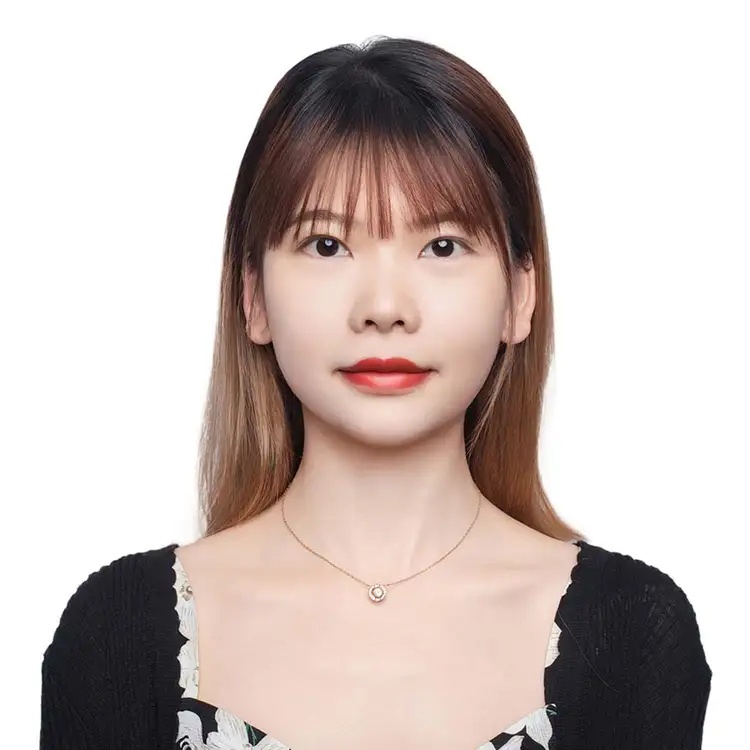}}]{Lijun Zhou}
received her Ph.D.\ degree from the University of Chinese Academy of Sciences, Beijing, China, in 2021.
She is currently an Algorithm Researcher at Xiaomi EV.
Her research interests include world models, video generation, and autonomous driving.
\end{IEEEbiography}
\begin{IEEEbiography}[{\biosphoto{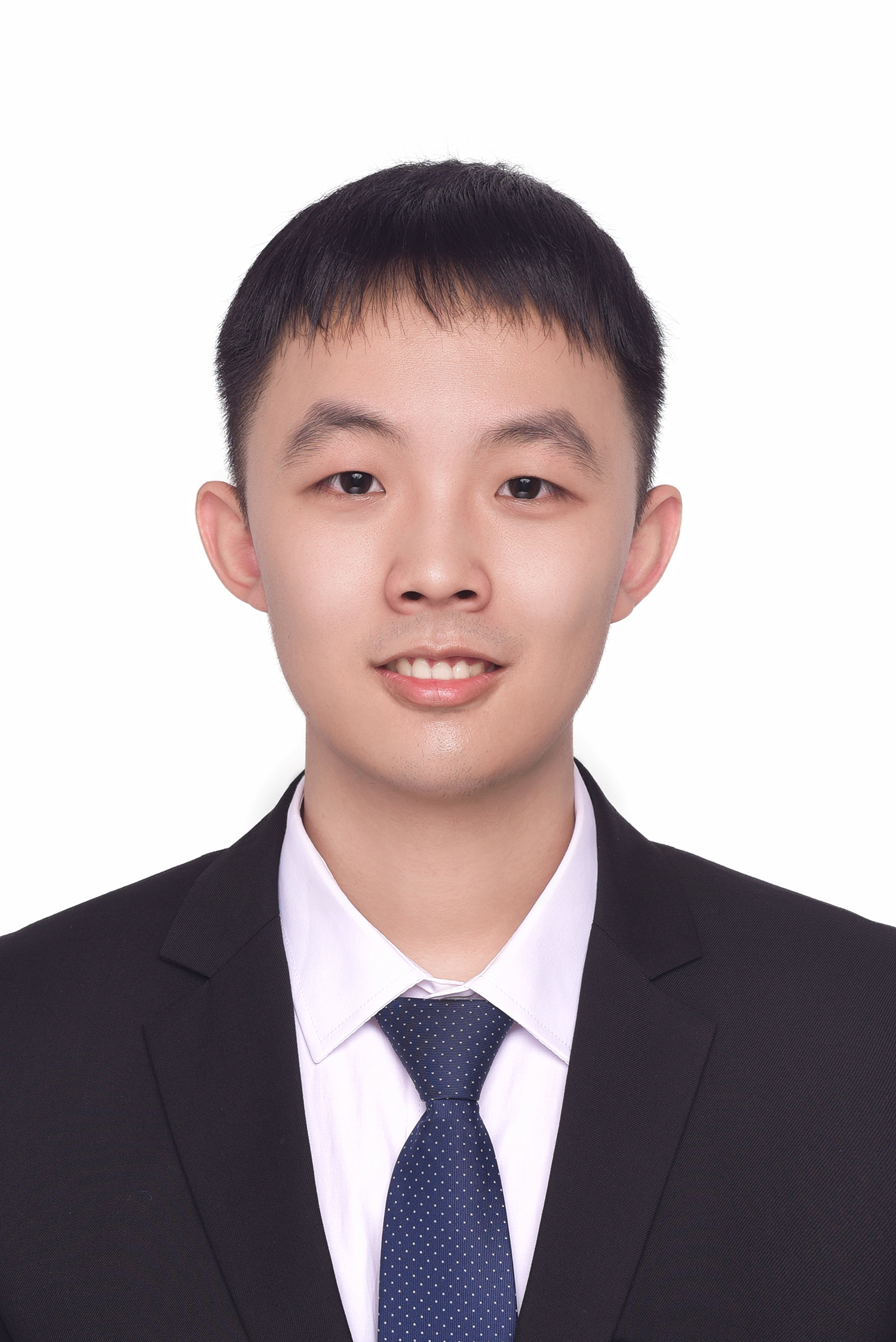}}]{Kaixin Xiong}
received the master's degree in information and communication engineering from the Huazhong University of Science and Technology, Wuhan, China, in 2023.
He is currently an Algorithm Engineer at Xiaomi EV.
His research interests include world models, 3D vision, and autonomous driving.
\end{IEEEbiography}
\begin{IEEEbiography}[{\biosphoto{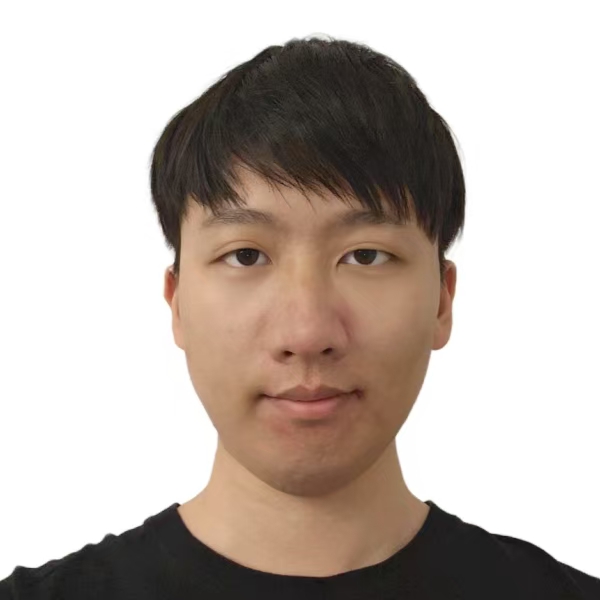}}]{Jingfeng Yao}
is a Ph.D.\ student with the Department of Electronic Information and Communications, Huazhong University of Science and Technology.
He is supervised by Prof.\ Xinggang Wang and is expected to graduate in 2027.
His previous research includes image matting and representation learning.
His current research interests focus on generative models.
\end{IEEEbiography}
\begin{IEEEbiography}[{\biosphoto{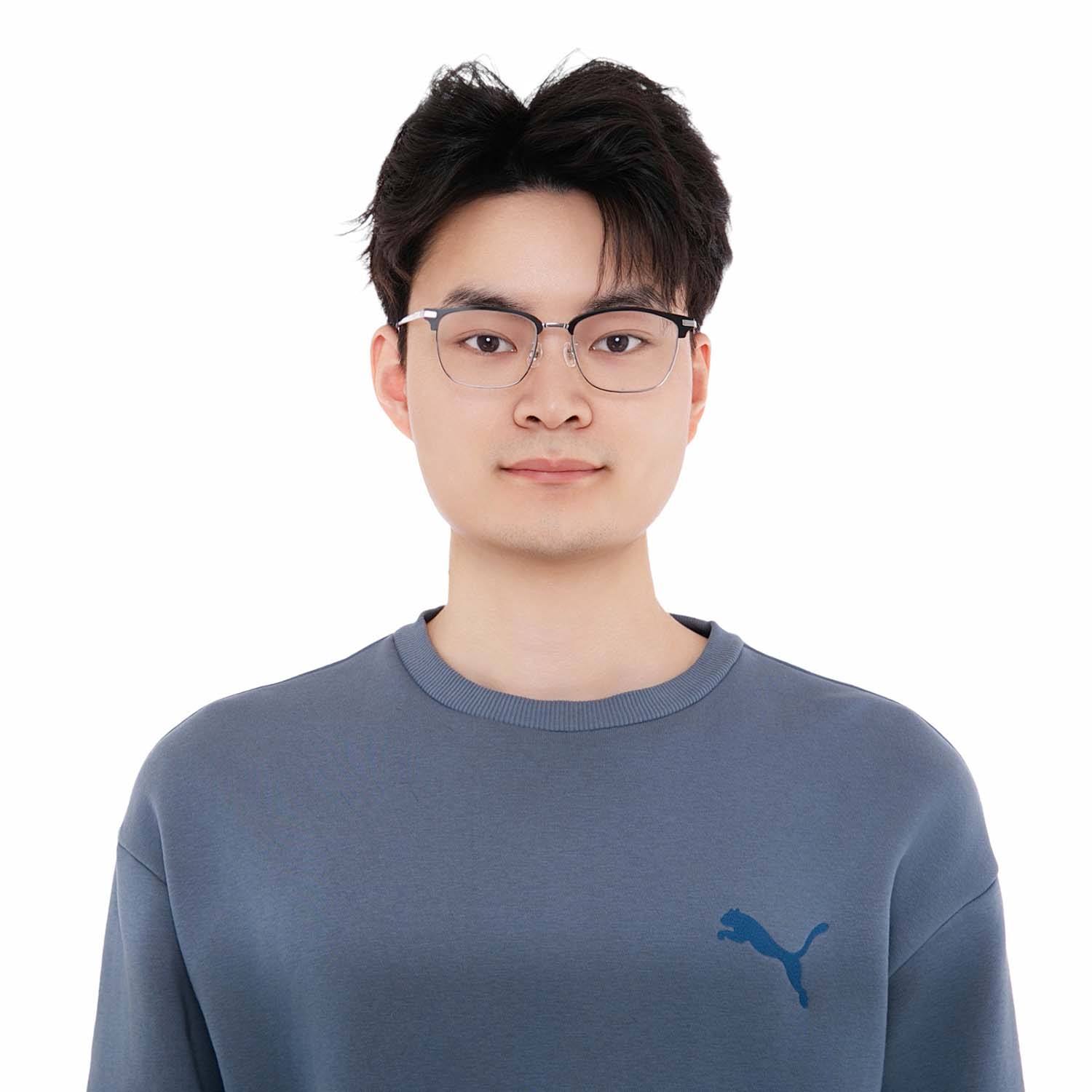}}]{Zhenxin Zhu}
received the M.S.\ degree in Control Science and Engineering from Beihang University, Beijing, China, in 2025.
He is currently an Algorithm Engineer at Xiaomi EV.
His research interests include physical AI and world models.
\end{IEEEbiography}
\begin{IEEEbiography}[{\biosphoto{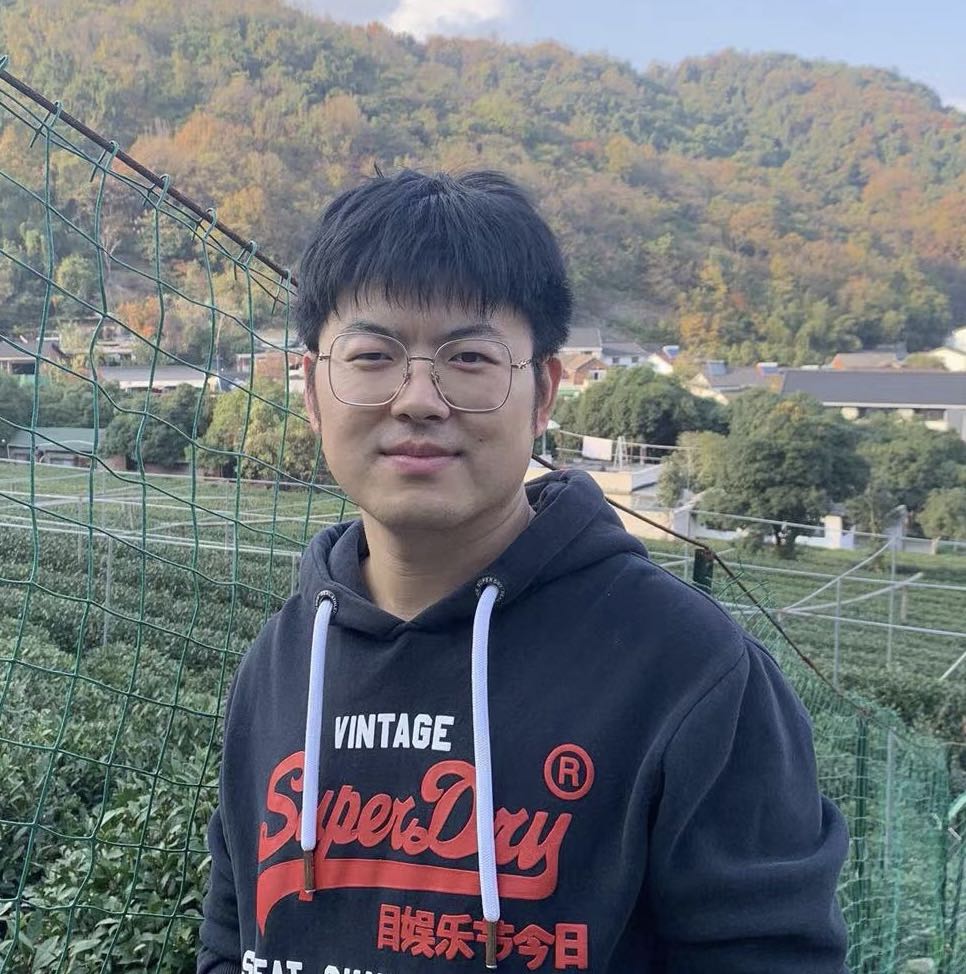}}]{Haiyang Sun}
received the master's degree in information and communication engineering from Tsinghua University, Beijing, China, in 2016.
He is currently an Expert Algorithm Engineer at Xiaomi EV.
His research interests include world models, 3D vision, and autonomous driving.
\end{IEEEbiography}
\begin{IEEEbiography}[{\biosphoto{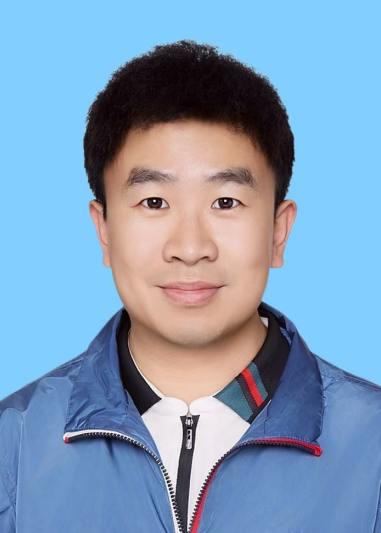}}]{Bing Wang}
received his Ph.D.\ degree from the School of Electrical and Electronic Engineering, Nanyang Technological University, Singapore, in 2016.
He is currently an Expert Algorithm Engineer at Xiaomi EV.
His research interests include computer vision, machine learning, world models, autonomous driving, and robotics.
\end{IEEEbiography}
\begin{IEEEbiography}[{\biosphoto{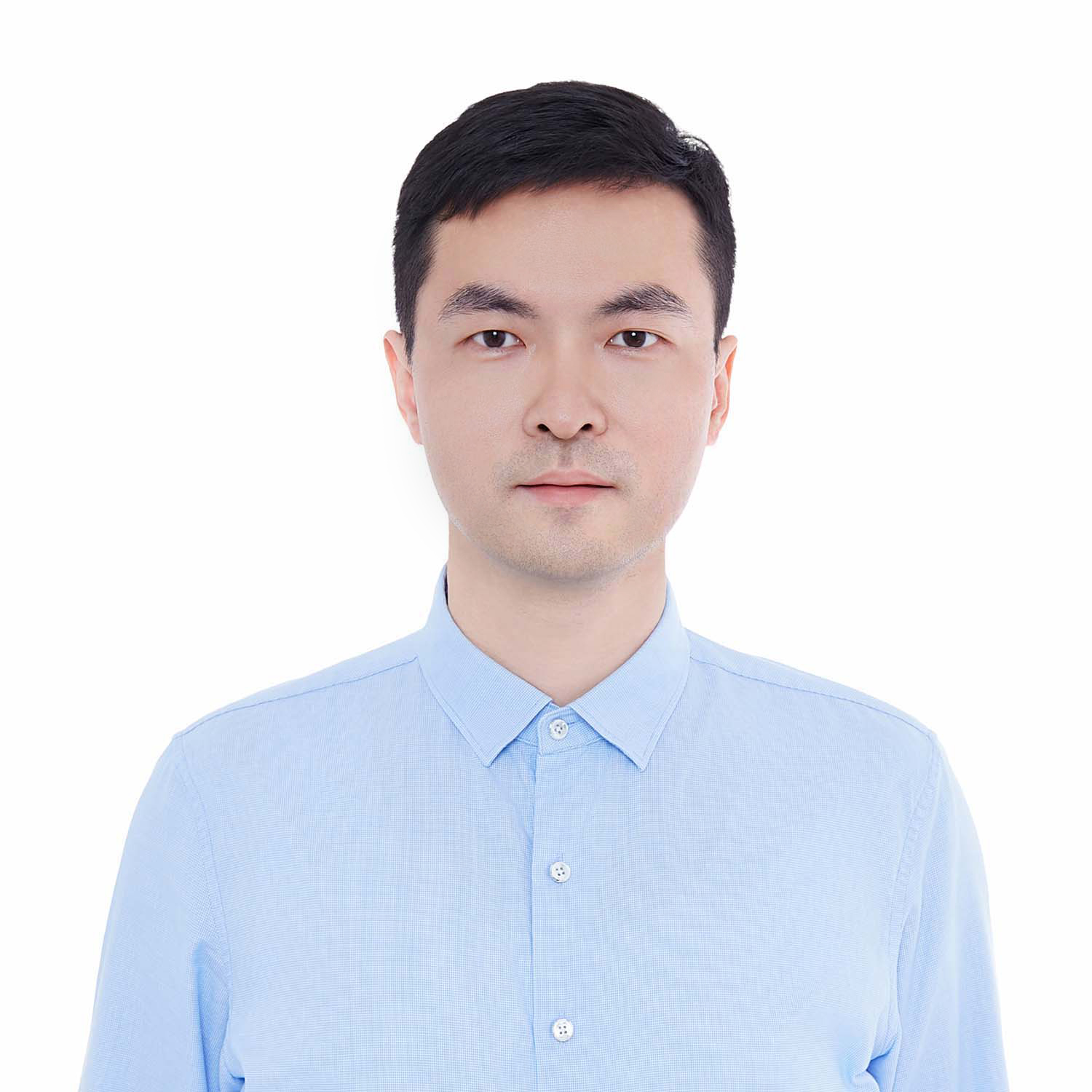}}]{Guang Chen}
received the Ph.D.\ degree from the Electrical and Computer Department of the University of Missouri in 2014.
He is currently an Expert Algorithm Engineer at Xiaomi EV.
His research interests include computer vision, machine learning, and autonomous driving.
\end{IEEEbiography}
\begin{IEEEbiography}[{\biosphoto{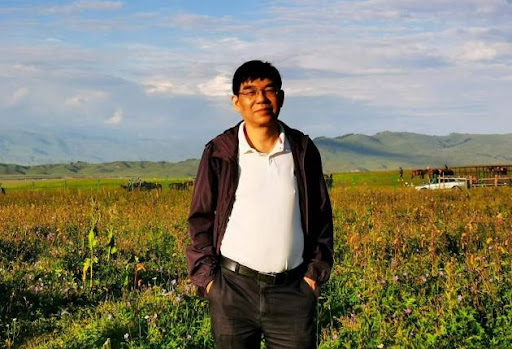}}]{Wenyu Liu (SM'15)}
received the B.S.\ degree in computer science from Tsinghua University, Beijing, China, in 1986, and the M.S.\ and Ph.D.\ degrees in electronics and information engineering from Huazhong University of Science and Technology (HUST), Wuhan, China, in 1991 and 2001, respectively.
He is currently a Professor with the School of Electronic Information and Communications, HUST.
His research interests include computer vision, multimedia, and machine learning.
\end{IEEEbiography}
\begin{IEEEbiography}[{\biosphoto{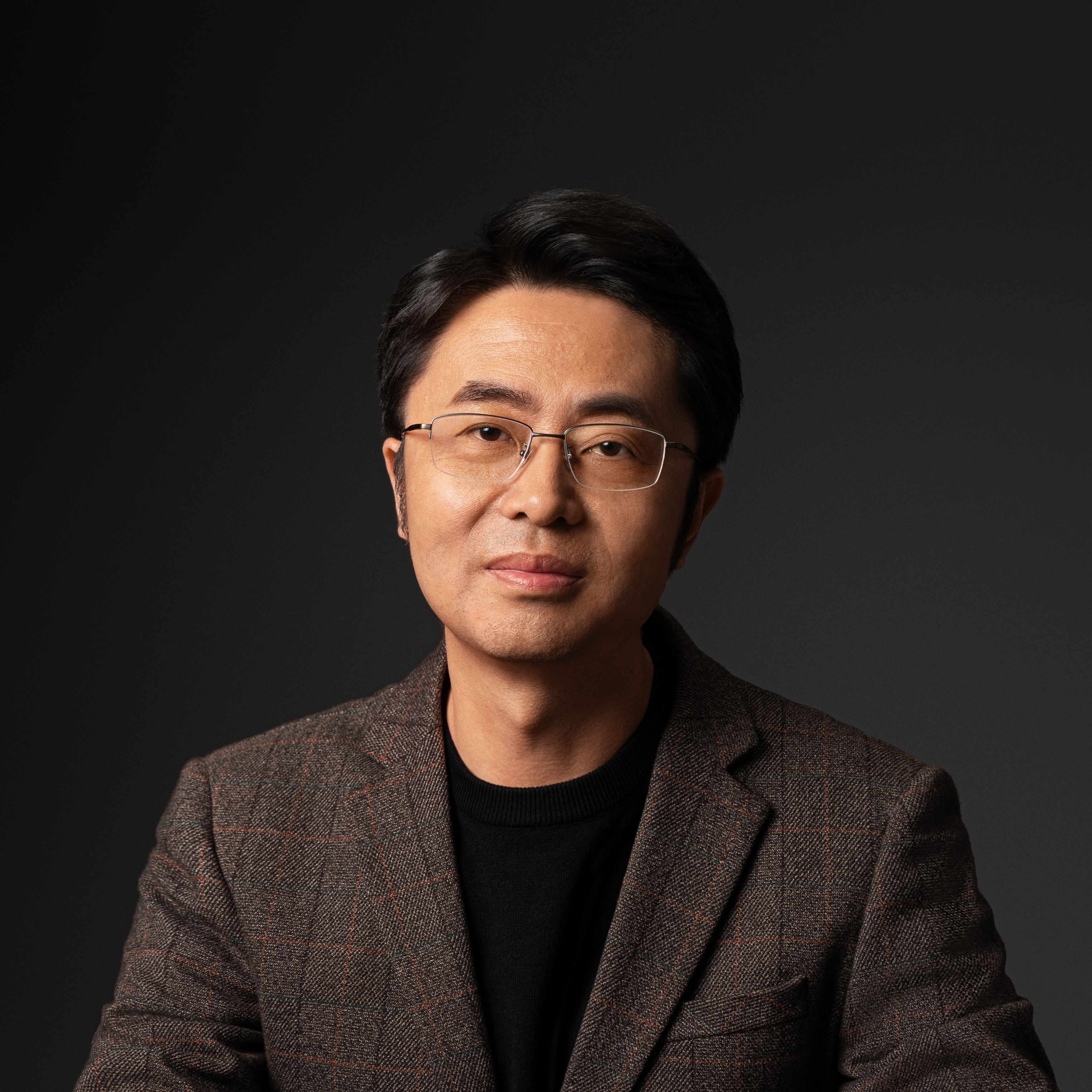}}]{Hangjun Ye}
received his Ph.D.\ degree from the Department of Computer Science and Technology, Tsinghua University, China, in 2003.
He is currently the head of the Autonomous Driving and Robotics Division, Xiaomi EV.
His research interests include computer vision, machine learning, autonomous driving, and robotics.
\end{IEEEbiography}
\begin{IEEEbiography}[{\biosphoto{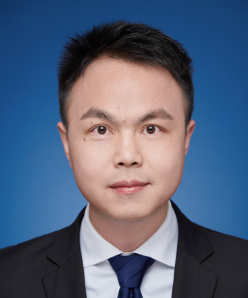}}]{Xinggang Wang (SM'24)}
received the B.S.\ and Ph.D.\ degrees in electronics and information engineering from Huazhong University of Science and Technology (HUST), Wuhan, China, in 2009 and 2014, respectively.
He was a Visiting Scholar with Temple University and the University of California, Los Angeles.
He is currently a Professor with the School of Electronic Information and Communications, HUST.
His research interests include visual representation learning, visual perception, visual planning, and multimodal foundation models.
He has more than 60{,}000 Google Scholar citations and an h-index of 90.
He is the Co-Editor-in-Chief of \emph{Image and Vision Computing}, an Associate Editor of \emph{IEEE TPAMI} and \emph{Machine Vision Applications}, and has served as an Area Chair for CVPR, ICCV, and NeurIPS.
\end{IEEEbiography}

\end{document}